%% file: main.tex
\documentclass[10pt,twocolumn,letterpaper]{article}

\usepackage[pagenumbers]{iccv} %

\input{preamble}

\usepackage{graphicx} %
\usepackage{booktabs} %
\usepackage{pifont}
\usepackage{caption} %
\definecolor{iccvblue}{rgb}{0.21,0.49,0.74}
\usepackage[pagebackref,breaklinks,colorlinks,allcolors=iccvblue]{hyperref}
\usepackage{footnote} %
\usepackage{multirow}
\makesavenoteenv{table} %
\makesavenoteenv{figure} %

\showeditstrue

\author{
Loïck Chambon$^{1,2}$ \quad
Eloi Zablocki$^{2}$ \quad
Alexandre Boulch$^{2}$ \quad
Mickaël Chen$^{3}$ \quad
Matthieu Cord$^{1,2}$ \quad
\and
\and
\large
\hspace{-3mm} \textsuperscript{1}ValeoAI, Paris, France.
\hspace{1mm} \textsuperscript{2}Sorbonne University, Paris, France.
\hspace{1mm} \textsuperscript{3}Hcompany.ai, Paris, France.
}

\title{GaussRender: Learning 3D Occupancy with Gaussian Rendering}

\begin{document}
\maketitle
\input{sec/0_abstract}    
\input{sec/1_intro}

\input{sec/2_related_work}

\input{sec/3_method}

\input{sec/4_experiments}

\input{sec/5_conclusion}

{
    \small
    \bibliographystyle{ieeenat_fullname}
    \bibliography{main}
}

\input{sec/X_suppl}

\end{document}

%% file: preamble.tex
\newif\ifshowedits

\newcommand{\addeditor}[3]{%
  \definecolor{#1color}{rgb}{#3}
  \expandafter\newcommand\csname #1\endcsname[1]{%
  \ifshowedits
    {\color{#1color} ##1}%
  \else
    {##1}%
  \fi
  }%
  \expandafter\newcommand\csname #1rmk\endcsname[1]{%
  \ifshowedits
    {\color{#1color} {\bf [#2: ##1]}}
  \fi
  }%
  \expandafter\newcommand\csname #1rpl\endcsname[2]{%
  \ifshowedits
    {\color{#1color} ##1 \sout{##2}}
  \else
    {##1}
  \fi
  }%
}

\newcommand{\method}{GaussRender}

\usepackage{pgfplots}
\pgfplotsset{compat=1.17}

\usepackage{colortbl}
\usepackage{booktabs}

\definecolor{nbarrier}{RGB}{255, 120, 50}
\definecolor{nbicycle}{RGB}{255, 192, 203}
\definecolor{nbus}{RGB}{255, 255, 0}
\definecolor{ncar}{RGB}{0, 150, 245}
\definecolor{nconstruct}{RGB}{0, 255, 255}
\definecolor{nmotor}{RGB}{200, 180, 0}
\definecolor{npedestrian}{RGB}{255, 0, 0}
\definecolor{ntraffic}{RGB}{255, 240, 150}
\definecolor{ntrailer}{RGB}{135, 60, 0}
\definecolor{ntruck}{RGB}{160, 32, 240}
\definecolor{ndriveable}{RGB}{255, 0, 255}
\definecolor{nother}{RGB}{139, 137, 137}
\definecolor{nsidewalk}{RGB}{75, 0, 75}
\definecolor{nterrain}{RGB}{150, 240, 80}
\definecolor{nmanmade}{RGB}{230, 230, 250}
\definecolor{nvegetation}{RGB}{0, 175, 0}
\definecolor{nothers}{RGB}{0, 0, 0}

\definecolor{col1}{RGB}{232, 161, 148}
\definecolor{col2}{RGB}{148, 187, 232}
\newcommand{\cmark}{\ding{51}}
\newcommand{\xmark}{\ding{55}}

\usepackage{makecell}

%% file: sec/0_abstract.tex
\begin{abstract}
Understanding the 3D geometry and semantics of driving scenes is critical for safe autonomous driving. Recent advances in 3D occupancy prediction have improved scene representation but often suffer from visual inconsistencies, leading to floating artifacts and poor surface localization. Existing voxel-wise losses (e.g., cross-entropy) fail to enforce visible geometric coherence.
In this paper, we propose \method{}, a module that improves 3D occupancy learning by enforcing projective consistency.  Our key idea is to project both predicted and ground-truth 3D occupancy into 2D camera views, where we apply supervision. Our method penalizes 3D configurations that produce inconsistent 2D projections, thereby enforcing a more coherent 3D structure.  To achieve this efficiently, we leverage differentiable rendering with Gaussian splatting. \method{} seamlessly integrates with existing architectures while maintaining efficiency and requiring no inference-time modifications. 
Extensive evaluations on multiple benchmarks (SurroundOcc-nuScenes, Occ3D-nuScenes, SSCBench-KITTI360) demonstrate that \method{} significantly improves geometric fidelity across various 3D occupancy models (TPVFormer, SurroundOcc, Symphonies), achieving state-of-the-art results, particularly on surface-sensitive metrics such as RayIoU.
The code is open-sourced at \url{https://github.com/valeoai/GaussRender}.
\end{abstract}

\begin{figure}
    \centering
    \includegraphics[width=\linewidth]{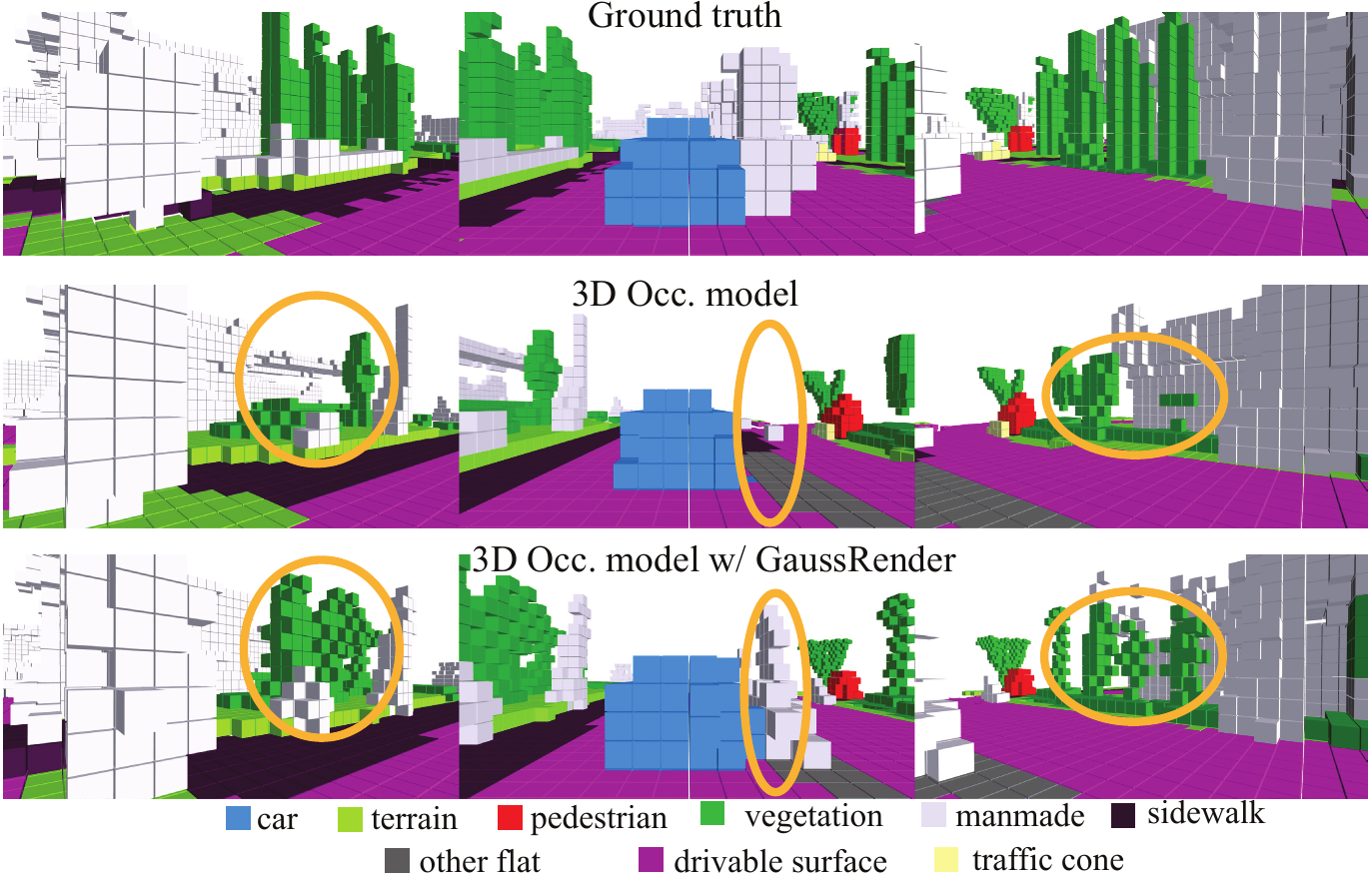}
    \caption{\textbf{Comparison of rendered 3D predictions.} Standard 3D Occupancy models trained on per-voxel losses result in physically implausible predictions (e.g., floating voxels, poorly localized surfaces, highlighted with \textcolor{orange}{orange} ellipses) that maintain high 3D IoU but fail to produce visually consistent predictions. \method{} enforces multi-view consistency, eliminating artifacts through learning projective constraints.}
    \label{fig:teaser_quali}
\end{figure}

%% file: sec/1_intro.tex
\section{Introduction}

Understanding the 3D geometry and semantics of driving scenes from multiple cameras is both a fundamental challenge and a critical requirement for autonomous driving. This problem is central to perception tasks such as object detection \cite{li2022bevformer, liu2022petr, li2023bevdepth, lin2022sparse4d, jiang2024far3d, zou2024unim}, agent forecasting \cite{kim2021lapred, yuan2021agentformer, gu2023vip3d, feng2024unitraj, shi2024mtrpp, shi2022motion, xu2024towardsmotion, xu2024ppt}, and scene segmentation \cite{chambon2024pointbev, gui2024fiptr, hu2021fiery, hu2022stp3, bartoccioni2022lara, harley2022simple, sirkogalouchenko2024occfeat}.
3D occupancy prediction \cite{vobecky2023pop3d, huang2023tpv, wei2023surroundocc, jiang2024symphonies, li2023fbocc3d} has emerged as a task to evaluate how well models capture the spatial structure and semantics of a scene.

The challenge in 3D occupancy prediction lies in achieving geometrically coherent reasoning from multi-view images.
Existing methods \citep{wei2023surroundocc,huang2023tpv,huang2024gaussianformer2,li2023voxformer} typically optimize per-voxel predictions using standard 3D losses, such as cross-entropy, Dice \cite{sudre2017generalised}, or Lovász \cite{berman2018lovasz}. However, these losses treat all voxels equally and do not enforce spatial consistency between neighboring voxels, leading to visual artifacts, as noticed in prior works \citep{liu2024sparseocc} and illustrated in \autoref{fig:teaser_quali}. 
Such artifacts (floating voxels, disjoint surfaces, and misaligned boundaries) have a low impact on voxel-based segmentation losses, which give the same weight to all voxels.
The consequences extend beyond mere visual artifacts: poor surface localization and unrealistic floating objects may significantly undermine downstream tasks such as free-space estimation or motion planning.
The underlying issue is that conventional supervision methods lack mechanisms to penalize unrealistic spatial arrangements, a deficiency that becomes especially apparent when projecting 3D artifacts into 2D views.
This observation motivates our key insight: integrating projective consistency into the training objective encourages the model to learn consistent, physically and visually plausible geometries.

We introduce \method{}, a module that bridges the gap through differentiable rendering of 3D occupancy predictions.
Our key idea is to enforce spatial consistency by projecting both predicted and ground-truth voxels into 2D camera views using Gaussian splatting \cite{kerbl3Dgaussians}.
These projections serve, during training, as a supervision signal, allowing us to penalize 3D configurations that produce poor 2D projections, thereby enforcing a more coherent and geometrically plausible 3D structure.
To achieve this, we apply two complementary supervision signals:
(1) a semantic rendering loss that enforces local semantic coherence, 
and (2) a depth rendering loss that penalizes occlusion-disrupting artifacts.
These rendering-based losses are applied alongside standard 3D supervision, such as cross-entropy, ensuring compatibility with existing occupancy learning frameworks.
{\method{} enables rendering from arbitrary viewpoints.}
Its flexibility helps reduce occlusions, for instance, by leveraging elevated viewpoints 
that are less affected by horizontal obstructions from ground objects.

We validate our approach on multiple datasets, including SurroundOcc-nuScenes \cite{wei2023surroundocc}, Occ3D-nuScenes \cite{tian2023occ3d}, and SSCBench-KITTI360 \cite{li2024sscbench}. Our experiments demonstrate that \method{} significantly improves geometric fidelity across diverse architectures --- ranging from multi-scale voxel-based models (SurroundOcc \cite{wei2023surroundocc}), tri-plane models (TPVFormer \cite{huang2023tpv}) to hybrid query-voxel models (Symphonies \cite{jiang2024symphonies}).
Across all settings and datasets, \method{} consistently improves performance on classical metrics such as IoU and mIoU. Moreover, when evaluated using surface- and artifact-sensitive metrics, such as RayIoU~\cite{liu2024sparseocc},
the improvements are even more pronounced. This is because the projective constraints enforced by our method promote surface continuity --- ensuring that neighboring voxels agree when rendered from any viewpoint, thereby eliminating floating artifacts and discontinuities.
Crucially, our approach is plug-and-play, integrating seamlessly with existing 3D occupancy frameworks without requiring any architectural modifications. Thanks to our efficient Gaussian rendering proxy, \method{} incurs minimal computational and memory overhead compared to NeRF-based alternatives \cite{huang2024selfocc, pan2024renderocc}.

The key contributions include:
\begin{itemize}
    \item {A rendering-based {module} that enforces semantic and geometric consistency in 3D occupancy prediction, eliminating visual and spatial artifacts through projective constraints,}
    \item {A fast and memory-efficient loss implementation-based on Gaussian splatting} without architectural modifications during training and having no impact during inference,
    \item {A camera positioning strategy} that amplifies supervision signals in geometrically complex regions.
    \item Improves results on three standard benchmarks for many models, with particular gains in RayIoU.
\end{itemize}

%% file: sec/2_related_work.tex
\section{Related work}
\label{sec:rw}

\subsection{Learning 3D semantic geometry from cameras}

Reconstructing unified 3D scenes from multi-camera systems must address three interconnected challenges: multi-view cameras create perspective limitations requiring occlusion reasoning, lifting 2D image features to metric 3D space demands geometric and semantic 2D-to-3D transfer, and merging multi-view inputs into volumetric representations like voxels incurs cubic memory growth \citep{xu2023survey3Docc}.
Modern methods address these challenges through structured intermediate representations.

Discrete methods rely on regular 3D grids to stay as close as possible to the desired voxel output.
Bird’s-eye-view (BeV) projections \cite{tian2024mambaocc, li2024viewformer} significantly reduce memory cost but introduce a substantial bias in the intermediate representation due to height compression.
Instead, tri-plane features \cite{huang2023tpv} and tensor decompositions \cite{zhao2024lowrankocc} map any 3D coordinate onto distinct planes or vectors performing interpolation to get distinct features. As opposed to BeV approaches, they do not suffer from height compression while still operating on a compressed representation.
Octrees \cite{lu2023octreeocc} preserves a full multi-scale 3D representation, coarse in uniform areas and fine where details are needed, resulting in a fast and memory-efficient representation~\cite{wei2023surroundocc}.
Continuous architectures have emerged, using a set of Gaussians to represent the scene \cite{huang2024gaussianformer2, huang2024gaussian} which {are then} discretized to obtain the voxelized output map. Furthermore, to better capture the overall instance semantics and scene context, query-based methods have been introduced \cite{jiang2024symphonies}. They facilitate interactions between image and volume features using sparse instance queries.

\begin{figure*}[t]
\centering
    \includegraphics[width=1.0\linewidth]{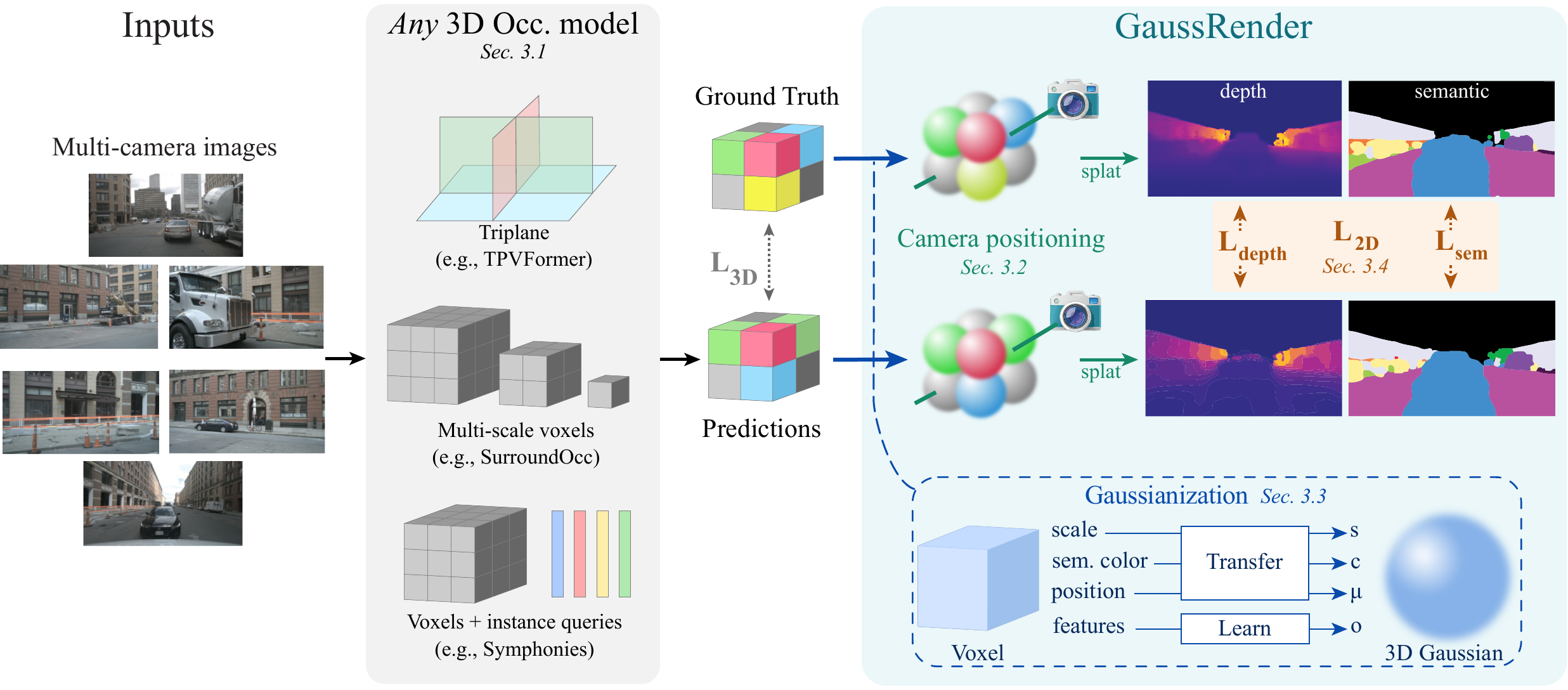}
    \caption{\textbf{Overview of \method{}.}
    Our module enforces 3D-2D consistency via differentiable Gaussian rendering. First, both predicted and ground-truth voxel grids are `gaussianized' by converting each voxel into a simple spherical Gaussian: the center $\mu$ is fixed at the voxel center, {the scale $s$ is a simple fixed scaling of the original voxel dimensions}, and features are directly transferred {as the semantic `classes' $c$} --- with only the opacity $o$ learned when voxel features are available. Next, virtual cameras are positioned in the scene (a fixed bird’s-eye view and a dynamic, arbitrarily placed camera as described in \autoref{sec:methods:camera}). The resulting 3D Gaussians are then projected into 2D using Gaussian splatting (\autoref{sec:methods:gaussian_rendering}), producing both semantic and depth renderings. These rendered views are compared against their ground-truth counterparts using an L1 penalty, ensuring enhanced spatial coherence and geometric consistency (\autoref{sec:methods:l2d_details}).
    }
    \label{fig:pipeline}
\end{figure*}

Complementary approaches have been proposed to enhance 3D occupancy predictions beyond architectural choices.
Temporal aggregation uses past frames to resolve occlusions and refine geometric details \cite{li2024viewformer, ye2024cvtoccc, li2024hierarchical, silva2024uniftri, pan2024renderocc}, with extensions to 4D forecasting for dynamic scenes \cite{khurana2023point, yang2025drivingoccupancy, wang2024occsora, ma2024cotr}.
Self-supervised methods reduces dependency on 3D annotations by generating pseudo-labels from monocular depth and segmentation \cite{huang2024selfocc, gan2024gaussianocc}, but suffer from scale miscalibrations and label mismatches between 2D and 3D labels.
Lastly, supervised rendering with lidar reprojections \cite{pan2024renderocc,sun2024gsrender} has been introduced to provide accurate depths. While more precise than pseudo-annotations, lidar reprojections face challenges such as signal sparsity, occlusions, and misalignment between the lidar and cameras.
{In particular, \cite{pan2024renderocc,sun2024gsrender} require temporal supervision from adjacent frames to compensate for lidar sparsity and occlusions. In contrast, our method achieves accurate supervision without these constraints, allowing more flexible and efficient 3D occupancy learning across different architectures.
}

\input{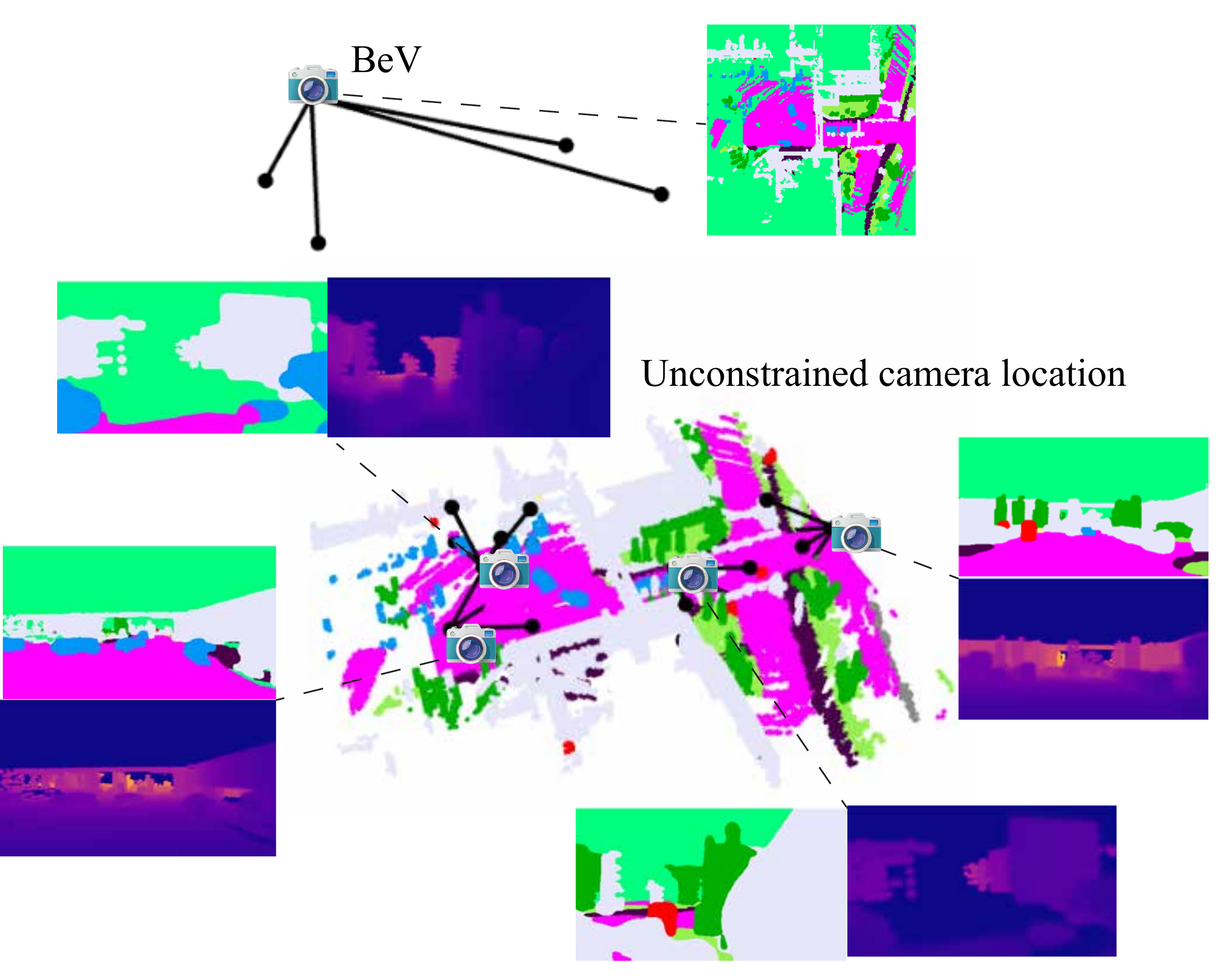}

Despite architectural variations, all methods ultimately produce voxel grids for supervision. This common 
output space lets \method{} enhance any 3D occupancy model with a simple rendering loss applied to the voxelic output.

\subsection{Differentiable rendering of 3D representations}

Supervising 3D predictions through 2D projections requires efficient differentiable rendering methods.
While traditional differentiable rendering methods handle 3D modalities like point clouds and meshes \cite{kato2018meshrenderer, miu2018paparazzi}, recent approaches focus on neural rendering and Gaussian-based methods.

NeRF-based methods \cite{mildenhall2020NeRF} perform volume rendering by predicting a volumetric density and applying ray-based integration. RenderOcc \cite{pan2024renderocc} constructs a NeRF-style volume representation supervised by semantic LiDAR projections. It leverages lidar-derived 2D labels (depth and semantics) for 3D supervision but inherits lidar's sparsity.
SelfOcc \cite{huang2024selfocc} uses signed distance fields for occupancy prediction and applies differentiable volume rendering to synthesize depth and semantic views. Self-supervision is enforced using multi-view consistency from video sequences.
However, NeRF-based rendering is computationally expensive, especially with high ray-sampling resolutions. Additionally, its reliance on image quality and occlusion mitigation necessitates auxiliary supervision from temporal frames \citep{sun2024gsrender,pan2024renderocc}.

Gaussian splatting provides an efficient alternative to NeRF by representing 3D scenes as a set of Gaussians \citep{kerbl3Dgaussians}. In 3D semantic occupancy prediction, GaussianOcc \cite{gan2024gaussianocc} projects a voxel-based Gaussian representation camera, using adjacent views for self-supervision, with optional 2D segmentation to refine the predicted occupancy.
{GSRender~\cite{sun2024gsrender} uses lidar reprojection and temporal consistency to help the learning process.}
GaussTR \cite{jiang2024gausstr} extends Gaussian representation learning with foundation models (e.g., CLIP, Grounded-SAM) to enable open-vocabulary occupancy prediction. Unlike other self-supervised methods, it does not use temporal supervision but relies on heavy pre-trained vision models.
A key limitation of these approaches is their tight coupling of Gaussian representations with the model architecture, imposing a fixed representation that limits flexibility. 
In contrast, our approach introduces a Gaussian rendering loss applicable to any model, without requiring the underlying 3D representation to be Gaussian.

All aforementioned methods are restricted to rendering from recorded camera perspectives, as they require RGB images to estimate pseudo-labels \citep{huang2024selfocc,gan2024gaussianocc,jiang2024gausstr} or rely on semantic LiDAR reprojections \citep{pan2024renderocc,sun2024gsrender}.
{In contrast, as \method{} renders both the ground truth and the predictions, it is not constrained by fixed camera perspectives or temporal supervision, enabling rendering from any viewpoint.}
Moreover, \method{} does not enforce a specific 3D feature representation, {and operates at prediction-level}, making it adaptable to diverse architectures.

%% file: fig/camera_location.tex
\begin{figure*}
    \centering
    \small
    \setlength{\tabcolsep}{16pt}
    \begin{tabular}{ccc}
         \includegraphics[width=0.24\linewidth]{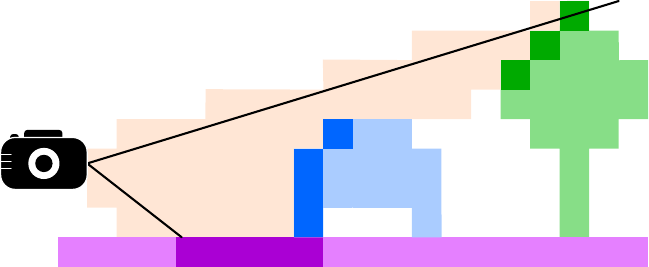}&
         \includegraphics[width=0.24\linewidth]{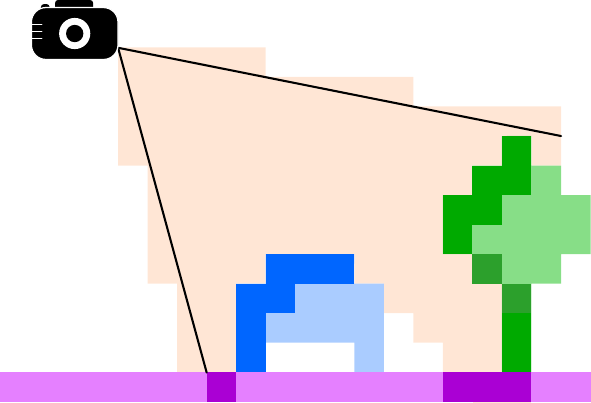}&
         \includegraphics[width=0.24\linewidth]{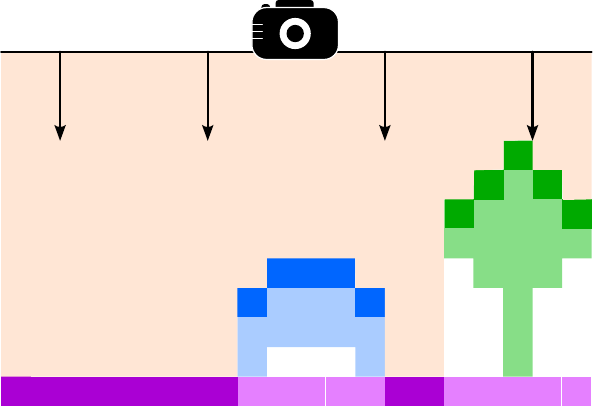} \\
         (a) Sensor view &
         (b) Virtual elevated view &
         (c) Virtual Orthographic BeV
    \end{tabular}
    \caption{\textbf{Arbitrary camera positioning.} Virtual cameras can be freely placed in the scene, enabling view constraints in occluded regions. The sensors' cameras do not provide access to the rear of objects, leading to occlusion issues. \method{} can place the cameras arbitrarily within the scene, allowing them to be elevated to reach previously hidden areas.}
    \label{fig:camera_location}
\end{figure*}

%% file: sec/3_method.tex
\section{GaussRender}
\label{sec:methods}

We present \method{}, a plug-and-play rendering module that enhances 3D occupancy models through efficient, differentiable Gaussian rendering. First, we define 3D-2D consistency and how to enforce it during training (\autoref{sec:methods:losses}).
We then outline our camera placement strategy (\autoref{sec:methods:camera}) and describe the rendering process, which projects 3D voxels into 2D images using Gaussian splatting (\autoref{sec:methods:gaussian_rendering}).
Finally, we detail %
{our} 2D rendering loss (\autoref{sec:methods:l2d_details}).

\subsection{Enforcing 3D-2D consistency}
\label{sec:methods:losses}
Vision-to-3D semantic occupancy models \citep{wei2023surroundocc,huang2023tpv,jiang2024symphonies} take a set of $N$ images, $I = \{I_i\}_{i=1}^N$, and predict a 3D semantic grid $O \in [0,1]^{X \times Y \times Z \times C}$, where $C$ is the number of semantic classes and $(X, Y, Z)$ defines the spatial resolution.
The standard pipeline consists of three steps:
\begin{itemize}
    \item \textit{Feature Extraction:} Each image is processed by a backbone network to extract 2D features $F = \{F_i\}_{i=1}^N \in \mathbb{R}^{d_\textit{img}}$, where $d_\textit{img}$ is the feature dimension.
    \item \textit{3D Lifting:} The features are lifted into a 3D representation (e.g., voxels, tri-planes) using cross- and self-attention mechanisms.
    \item \textit{Voxel Prediction:} The 3D representation is converted into a voxel grid, and standard losses (e.g., cross-entropy, Lovász, Dice) are computed against the ground truth.
\end{itemize}

\paragraph{}
While the standard 3D losses (denoted as \(L_{\text{3D}}\)) ensure overall occupancy alignment {with the ground truth}, they treat each voxel independently and ignore spatial consistency. This limitation can lead to artifacts such as floating voxels or misaligned surfaces \citep{liu2024sparseocc}. To overcome these issues, we introduce a 2D rendering loss, \(L_{\text{2D}}\), which provides additional supervision by comparing rendered views of the predicted occupancy with corresponding ground-truth images. This loss enforces spatial coherence in the 2D projection, thereby penalizing 3D configurations that produce poor renderings. The overall training loss is defined as:
\begin{equation}
L = L_{\text{3D}} + \lambda L_{\text{2D}},
\end{equation}
where $\lambda \in \mathbb{R}^+$ is a weighting factor for the 2D loss.

In the following subsections, we define \(L_{\text{2D}}\) by addressing two key aspects: (1) the placement of rendering cameras in the scene (\autoref{sec:methods:camera}), and (2) the differentiable rendering of 3D occupancy via Gaussian splatting (\autoref{sec:methods:gaussian_rendering}).

\subsection{Camera Placement Strategy}
\label{sec:methods:camera}
\method{} is not restricted to the original sensor or adjacent frames \citep{pan2024renderocc, huang2024selfocc, sun2024gsrender}, it renders images from arbitrary camera positions within the scene. This flexibility allows us to position virtual cameras anywhere in the 3D space.
\autoref{fig:camera_location} illustrates {how virtual cameras generate complementary constraints (both semantic and depth) to enhance voxel consistency.}

The placement of these virtual cameras is crucial and has been extensively studied. Several strategies have been found depending on the objective:
\begin{itemize}
    \item \textbf{Visible Voxels:} If the goal is to accurately reconstruct only the visible portions of the scene \citep{tian2023occ3d, li2024sscbench}, the rendering cameras should be placed close to the original sensor positions{, as illustrated in \autoref{fig:camera_location} (a)} in order to constraint the visibile voxels. This placement ensures that the inferred 3D structure remains consistent with the sensor's view. 
    \item \textbf{Holistic 3D Reconstruction:} For complete 3D reconstruction, including occluded regions \citep{wei2023surroundocc}, cameras should be positioned more diversely. Exploring different viewpoints provides additional supervision for occluded areas, heavily penalizing aberrant configurations.
    {In practice, we generate the virtual views with a simple rule: the camera is 
    {(1) elevated along $z$ axis with respect to its original position, (2) randomly translated in a close range to the ego-vehicle on the $xy$ plane. 
    }
    The resulting camera increases the field of view, looking at both visible and occluded parts of the scene (\autoref{fig:camera_location} (b)).}
\end{itemize}

Additionally, given the importance of bird's-eye view (BeV) understanding in autonomous driving, we systematically include a fixed virtual {orthographic} BeV camera{, illustrated in \autoref{fig:camera_location} (c)}. The BeV camera offers an orthogonal and complementary perspective, ensuring that objects are accurately localized on the ground and further enhancing voxel consistency.

In practice, placing a camera involves specifying its extrinsic and intrinsic parameters, respectively defining the viewing transformation \(W \in \mathbb{R}^{4 \times 4}\) and the projective transformation \(K \in \mathbb{R}^{3 \times 3}\).
{Our method keeps the intrinsic parameters fixed and only update the extrinsic parameters for each training batch.
}

\subsection{Gaussian rendering}
\label{sec:methods:gaussian_rendering}

For each camera, once its position is chosen and its intrinsics and extrinsics are set, we render the 3D occupancy into the corresponding view. Our goal is fast, fully differentiable rendering that supports efficient gradient backpropagation. To do it, we adopt a Gaussian splatting approach \cite{kerbl3Dgaussians}, which is significantly faster than traditional ray-casting while preserving differentiability.
Consequently, we represent each voxel as a Gaussian primitive.

\paragraph{Voxel `Gaussianization'.}
To derive a Gaussian from a voxel, we emphasize simplicity to ease the learning process. In practice, we: (1) \textit{use spherical Gaussians:} represent each voxel as a sphere, which removes the need for orientation parameters; (2) \textit{fix the center:} place each Gaussian at the center of its corresponding voxel, eliminating the need to learn an offset; (3) \textit{fix the scale:} set the scale of each Gaussian based on the voxel dimensions.

Concretely, as illustrated in \autoref{fig:pipeline} (bottom right), for each voxel at position $\mu = (x,y,z)$, we create a simple Gaussian primitive with:  
\begin{itemize}
    \item position $\mu$: inherited from voxel grid coordinates.
    \item scale $S=\textit{Diag}(s)$: {a diagonal matrix defined by a scalar factor $s \in \mathbb{R}$.} %
    \item semantic `color' logits $c$: taken from the model's final prediction.
    \item opacity $o$: learned from voxel features (or derived from the logit of the empty semantic class when features are absent).
    \item rotation $R = I$: set to the identity matrix, as spheres require no orientation. With this choice, the Gaussian covariance matrix becomes $\Sigma_{3D}=S^2$.
\end{itemize}

\paragraph{Gaussian rendering.}
To relate the 3D Gaussian representation to the 2D image, we project the 3D covariance matrix \(\Sigma_{3D}\) into the image plane using the camera parameters. The projected covariance is given by:
\begin{equation}
    \Sigma_{2D} = J \cdot W \cdot \Sigma_{3D} \cdot W^T \cdot J^T,
\end{equation}

where \(J\) is the Jacobian of the affine approximation of the projective transformation \(K\) related to the intrinsic parameters (see \cite{kerbl3Dgaussians} for details).
This 2D covariance defines the shape and spread of each Gaussian in the image, which directly influences the computation of opacity and transmittance in the following rendering step.

{For each pixel $p$ in the 2D projection,} the rendering computes {its semantic value by aggregating contributions from all projected Gaussians.} Specifically, at $p$, the rendered semantic color value $C_p \in [0,1]^\mathcal{C}$ ($\mathcal{C}$ is the number of semantic classes) and the rendered depth $D_p \in \mathbb{R}^{+}$ are given by:
\begin{equation}
C_p = \sum_{i=1}^N T_i, \alpha_i, \mathbf{c}_i, \quad D_p = \sum_{i=1}^N T_i, \alpha_i, \mathbf{d}_i,
\label{eq:color_rendering}
\end{equation}

where:
\begin{itemize}
    \item \(N\) is the {total} number of Gaussians. %
    \item \(\alpha_i = 1 - \exp(-\sigma_i\,\delta_i)\) is the opacity (or alpha blending factor) for the \(i\)th Gaussian, with \(\sigma_i\) representing its density and \(\delta_i\) the distance traversed along the ray.
    \item \(T_i = \prod_{j=1}^{i-1} (1-\alpha_j)\) is the accumulated transmittance from all Gaussians closer to the camera, which accounts for occlusions.
    \item $\mathbf{c}_i {\in [0,1]^\mathcal{C}}$ is the `color' probability of the \(i\)th Gaussian.
    \item $\mathbf{d}_i {\in \mathbb{R}^+}$ is the distance of the \(i\)th Gaussian to the projected camera.
\end{itemize}

We apply the same rendering process to both the predicted semantic occupancy and the 3D ground truth. For ground-truth rendering, we render only occupied voxels by assigning them an opacity of 1 (and 0 for empty voxels), while for predictions we use the learned opacities.

This differentiable and efficient rendering pipeline thus translates the 3D occupancy into 2D views, enabling robust pixel-wise supervision to enforce spatial consistency.

\subsection{\texorpdfstring{{$L_{2D}$}}~~rendering loss computation}
\label{sec:methods:l2d_details}

{
For each virtual camera, note `*', and the associated pixel set $P^*$, we render semantic and depth images $I_{\text{sem}}^* = \{ C_p, p \in P^* \}$ and $ I_{\text{depth}}^*  = \{ D_p, p \in P^* \}$, from the predicted semantic occupancy, following \autoref{eq:color_rendering}. Similarly, we obtain \( \tilde{I}_{\text{sem}}^* \) and \( \tilde{I}_{\text{depth}}^* \) from the ground-truth voxels. 
}

To enforce consistency, we compare each predicted rendering against its ground-truth counterpart from the same viewpoint using the L1 distance:

\begin{equation}
    L_{\text{depth}}^* = \frac{1}{d_{\text{range}}^*} \, \| I_{\text{depth}}^{*} \text{-} \tilde{I}_{\text{depth}}^{*} \|_1,
    \,\,\,
    L_{\text{sem}}^{*} = \| I_{\text{sem}}^{*} \text{-} \tilde{I}_{\text{sem}}^{*} \|_1,
\end{equation}

where \( d_{\text{range}}^{*} \in \mathbb{R}^{+} \) is a normalization factor based on the maximum depth, ensuring scale consistency across different scenes.
The per-camera loss is then obtained as the sum of these two terms:
\begin{equation}
    L_{\text{2D}}^{*} = L_{\text{depth}}^{*} + L_{\text{sem}}^{*}.
\end{equation}

{
Thus, the overall 2D rendering loss for our module is:
\begin{equation}
L_{\text{2D}} =  L_{\text{2D}}^{\text{bev}} + L_{\text{2D}}^{\text{cam}}.
\label{eq:rendering_losses}
\end{equation}
where `$\text{bev}$' is the orthographic BeV camera (top-down perspective) crucial for global scene understanding and `$\text{cam}$' is the dynamically generated camera generated following the strategy outlined in \autoref{sec:methods:camera}, which ensures diverse viewpoints and improves generalization.
}

This loss ensures that the learned 3D occupancy aligns with both depth and semantic projections, improving the consistency between 3D and 2D representations.

%% file: sec/4_experiments.tex
\section{Experiments}
\label{sec:exp}

We evaluate \method{} on several models and datasets to demonstrate its versatility. We present experimental details in \autoref{sec:exp:details}, compare our results against state-of-the-art models and datasets in \autoref{sec:exp:sota}, show how \method{} enhances 3D semantic occupancy predictions from multiple views in \autoref{sec:exp:finer-details}, and present ablations in \autoref{sec:exp:ablations}.

\subsection{Data and models}
\label{sec:exp:details}

\paragraph{Data.}
The trainings and evaluations are conducted on three datasets: SurroundOcc-nuScenes \cite{wei2023surroundocc}, Occ3d-nuScenes \cite{tian2023occ3d}, and SSCBench-Kitti360 \cite{li2024sscbench}.
We briefly outline below the specific characteristics of each dataset. More details can be found in \autoref{app:datasets}.

\textbf{SurroundOcc-nuScenes} is derived from the nuScenes dataset \cite{caesar2020nuscenes}, acquired in Boston and Singapore. It aggregates the lidar annotations of nuScenes to create 3D semantic occupancy grids of range 
{$[\text{-}50,50] \! \times \! [\text{-}50,50] \! \times \! [\text{-}5,3]$ meters with 50cm voxel resolution}, 
with labels corresponding to 17 lidar semantic segmentation classes.
This dataset takes into account both visible and occluded voxels. The occluded voxels are obtained by accumulating lidar data over the frames of the whole sequence consequently introducing temporal artifacts over dynamic objects. 

\textbf{Occ3D-nuScenes} is also based on {the} nuScenes dataset. It contains 18 semantic classes and has 
{a 40cm voxel grid of range $[\text{-}40,40] \! \times \! [\text{-}40,40] \! \times \! [\text{-}1,5.4]$ meters.}
One major difference with SurroundOcc-nuScenes, is that it only evaluates the voxels visible from the cameras at the current time frame. Thus, it focuses on the geometric and semantic understanding of the visible objects, rather than extrapolating to occluded regions, leading to a simpler task.

\textbf{SSCBench-Kitti360} \cite{li2024sscbench} is derived from the Kitti360 dataset \cite{Liao2022PAMI}, acquired in Germany. 
{It contains 19 semantic classes and has a 20cm voxel grid of range $[0,51.2] \! \times \! [-25.6,25.6] \! \times \! [-2,4.4]$ meters, resulting in a very precise semantic of urban scenes.}
It evaluates both visible and occluded voxels, making the dataset particularly challenging due to its voxel resolution and the presence of occlusions.

\paragraph{Models and training details.}
We {integrate} \method{} into three different models that use different intermediate representations: \textbf{SurroundOcc} \cite{wei2023surroundocc} (multi-scale voxel-based approach), \textbf{TPVFormer} \cite{huang2023tpv} (tri-plane-based approach), and \textbf{Symphonies} \cite{jiang2024symphonies} (voxel-with-instance query-based approach).
By doing so we validate the claim that our proposed approach is compatible with any type of architecture.
For each combination of models and datasets, we follow the same procedure. {By default, we evaluate the original model checkpoints when available; otherwise, we report scores from previous papers if provided or re-train the models for the target dataset}. Each model uses the same training settings, following the optimization parameters of ~\cite{wei2023surroundocc}. 
{
Camera strategy parameters (elevation and translation are detailed in~\autoref{sec:exp:ablations}).
}
More {technical} details can be found in \autoref{app:models}.

\subsection{3D semantic occupancy results} %
\label{sec:exp:sota}

\input{tables/unified_benchmark}

We evaluate \method{} across multiple models \citep{wei2023surroundocc,huang2023tpv,jiang2024symphonies} and datasets \citep{wei2023surroundocc,tian2023occ3d,li2024sscbench} for 3D semantic occupancy prediction. Our method consistently improves performance without {relying on} other sensors such as the lidar used in other works \cite{pan2024renderocc, sun2024gsrender}. Results are summarized in \autoref{tab:unified_benchmark} and detailed scores by class can be found in \autoref{app:results}.

\textbf{SurroundOcc-nuScenes \citep{wei2023surroundocc}.}
On {this dataset}, considering visible and occluded voxels, \method{} brings significant gains to all tested models. As shown in \autoref{tab:unified_benchmark}, TPVFormer \cite{huang2023tpv} and SurroundOcc \cite{wei2023surroundocc} reach the top two ranks. They achieve higher IoU \textbf{(\textcolor{ForestGreen}{+1.2}} and \textbf{\textcolor{ForestGreen}{+1.1}}) and mIoU \textbf{(\textcolor{ForestGreen}{+3.8}} and \textbf{\textcolor{ForestGreen}{+0.5}}) compared to their original implementations.
Remarkably, \method{} enables these models to surpass {more} recent approaches like GaussianFormer and GaussianFormerV2 \cite{huang2024gaussianformer2} in both IoU and mIoU, proving that older architectures can achieve state-of-the-art results when {their training is enhanced with \method{}.}

\textbf{Occ3D-nuScenes.}
On Occ3D-nuScenes \cite{tian2023occ3d}, \method{} leads to the top two results: TPVFormer with \method{} achieves 30.48 mIoU (\textbf{\textcolor{ForestGreen}{+2.65}}), ranking first, and SurroundOcc with \method{} reaches 30.38 mIoU (\textbf{\textcolor{ForestGreen}{+1.17}}), ranking second.
Notably, our approach outperforms other rendering methods such as RenderOcc \cite{pan2024renderocc} and GSRender \cite{sun2024gsrender} without requiring lidar inputs for loss computation. In addition, we also evaluate our model using only the rendering losses and compare to other supervised and unsupervised methods, see\cref{tab:rendering_methods_comparison}. While being static (single frame), it outperforms both static (\textbf{\textcolor{ForestGreen}{+6.0}}$\,${\small mIoU}) and temporal (\textbf{\textcolor{ForestGreen}{+1.4}}$\,${\small mIoU}) versions of RenderOcc, the latter casting rays from other frames achieving a new state-of-the art score on 2D supervision for 3D Occupancy on Occ3d-nuScenes. This demonstrates {that methods augmented with \method{}} can learn strong 3D representations using only cameras. 

\input{tables/2D_training}

\textbf{SSCBench-Kitti360.}
On the challenging SSCBench-Kitti360 \cite{Liao2022PAMI} dataset, integrating \method{} to SurroundOcc \cite{wei2023surroundocc} and Symphonies \cite{jiang2024symphonies} {boosts} IoU (\textbf{\textcolor{ForestGreen}{+0.11}} and \textbf{\textcolor{ForestGreen}{+0.68}}) and mIoU (\textbf{\textcolor{ForestGreen}{+0.26}} and \textbf{\textcolor{ForestGreen}{+0.29}}).
Absolute gains are smaller because models in this benchmark achieve tightly clustered performance due to the smaller voxel size making the segmentation task more difficult. However, when we look only at the visible voxels or in the sensor image metrics, we better see the overall improvements as figured in \cref{tab:gaussrender_kitti360}. SurroundOcc with \method{} achieves higher mIoU (\textbf{\textcolor{ForestGreen}{+1.23}}) and IoU  (\textbf{\textcolor{ForestGreen}{+1.99}}) on visible voxels. It clearly appears that \method{} has strong impact on visible voxels meaning it has removed visibility artefacts.

\input{tables/kitti360_additional}

Our results show that \method{} consistently enhances various architectures leading to state-of-the-art results, reaches top results without requiring projected lidar annotations, and remains effective across different dataset scales and annotation densities. This demonstrates the significant advantages of \method{} for 3D semantic occupancy learning.

\subsection{Finer-grained multi-view metric analysis}
\label{sec:exp:finer-details}

\paragraph{RayIoU.}
\input{tables/rayiou}

{Classical 3D occupancy metrics, such as voxel-wise IoU, treat all voxels equally, often failing to capture inconsistencies in object surfaces and depth localization. This can lead to misleading evaluations, as models may artificially inflate scores by predicting thick or duplicated surfaces \citep{liu2024sparseocc} rather than accurately reconstructing scene geometry.}
{To address this, we use RayIoU \cite{liu2024sparseocc}, a metric designed to assess 3D occupancy predictions in a depth-aware manner. Instead of evaluating individual voxels, RayIoU casts rays through the predicted 3D volume and determines correctness based on the first occupied voxel the ray intersects. A prediction is considered correct if both the class label and depth fall within a given tolerance. This approach mitigates issues with voxel-level IoU, ensuring that models are rewarded for precise surface localization rather than over-segmentation.}

Using \method{} consistently improves RayIoU across tested architectures, as reported in \autoref{tab:rayiou_metrics}, highlighting its ability to enhance spatial consistency.
{Notably, models enhanced with \method{} achieve state-of-the-art performances with a single frame, outperforming prior works by a significant margin.} %

\subsection{Ablation studies}

{We conduct our ablations} on {fixed} subsets of the datasets{, each representing 20\% of the training and validation sets.} 

\label{sec:exp:ablations}

\input{fig/camera_strategy}

\subsubsection{Impact of supervising with virtual viewpoints}

A key advantage of \method{} is its ability to render 3D occupancy from arbitrary viewpoints, offering a flexible way to supervise 3D occupancy. To investigate the influence of virtual camera configurations, we evaluate five different camera placement strategies: Sensor, Elevated, Elevated + Around, Fully Random, Dynamic. More details in Appendix \ref{app:camera_strat}.
Quantitatively \autoref{tab:camera_sampling_strategies} quantifies the effect of each strategy under both 2D-only and 2D+3D supervision using mIoU. To do this we train a TPVFormer model with \method{} during 20\% of the Occ3d-nuSc training set. Several conclusions can be drawn. Under 2D-only supervision, the \textbf{Sensor strategy} leads by a large margin (16.1 mIoU), suggesting that when only 2D supervision is available, aligning virtual views with actual camera positions maximizes consistency and learning efficiency. Both \textbf{Fully Random} and \textbf{Dynamic} strategies suffer heavily, with performance dropping to 6.9 and 4.8 mIoU, respectively. These results highlight the risk of unstructured camera placement: random viewpoints often observe unoccupied or irrelevant parts of the scene, weakening the supervision signal while dynamic positioning may introduce instability and bad signals during training leading to poor results. Under 2D+3D supervision, we clearly see that what matters is to have complementary viewpoints since the Sensor strategy is beaten by \textbf{Elevated + Around} achieving the best result (26.3 mIoU). This indicates that providing diverse and informative viewpoints enhances learning, even when 3D supervision is available.

Overall, these findings emphasize that virtual camera placement is important. If we consider a 2D only training, positioning cameras at sensor location ensures consistency with sensors during training, while when training 2D+3D, we need additional viewpoints to enhance supervision.

\subsubsection{Loss increment}
Finally, we analyze the contribution of each individual loss component to the final metrics by gradually introducing each term. The results, in \autoref{tab:loss_components_study}, show that each successive addition of a loss component leads to a gradual improvement in performance, justifying the inclusion of each term.

\input{tables/loss_increment}

%% file: tables/unified_benchmark.tex
\begin{table}
    \centering
    \small
    \resizebox{\columnwidth}{!}
    {
    \setlength{\tabcolsep}{3pt}
    \begin{tabular}{@{}l c c c c c@{}}
        \toprule
              & \multicolumn{2}{c}{\textbf{Surround-}}  & \textbf{Occ3D} & \multicolumn{2}{c}{\textbf{SSCBench}}\\
              & \multicolumn{2}{c}{\textbf{Occ nusc.} \citep{wei2023surroundocc}}  & \textbf{nusc} \citep{tian2023occ3d} & \multicolumn{2}{c}{\textbf{KITTI360} \citep{li2024sscbench}}\\
              \textit{Evaluation split} & \multicolumn{2}{c}{\textit{val.}}  & \textit{val.} & \multicolumn{2}{c}{\textit{test}}\\
        Visible voxels only & \multicolumn{2}{c}{\ding{55}} & \ding{51} & \multicolumn{2}{c}{\ding{55}} \\
              \cmidrule(lr){2-3} \cmidrule(lr){4-4} \cmidrule(lr){5-6}
        Model &  IoU &  mIoU & mIoU & IoU &  mIoU\\
        \midrule
        BEVDet \citep{huang2021bevdet} & - & - & 19.38 & - & - \\
        BEVStereo \citep{li2023bevstereo} & - & - & 24.51 & - & - \\
        RenderOcc \citep{pan2024renderocc} & - & - & 26.11 & - & - \\
        CTF-Occ \citep{tian2023occ3d} & - & - & 28.53 & - & - \\
        GSRender \citep{sun2024gsrender} & - & - & 29.56 & - & - \\
        TPVFormer-lidar \citep{huang2023tpv} & 11.51 & 11.66 & -  & - & - \\
        MonoScene \cite{cao2022monoscene} & 23.96 & 7.31 & 6.06 & 37.87 & 12.31 \\
        Atlas \cite{murez2020atlas} & 28.66 & 15.00 & -  & - & - \\
        GaussianFormer \citep{huang2024gaussian} & 29.83 & {19.10} &  - & - & - \\
        BEVFormer \cite{li2022bevformer} & 30.50 & 16.75 & 26.88 & - & - \\
        GaussianFormerv2 \citep{huang2024gaussianformer2} & 30.56 & 20.02 & - & - & - \\
        VoxFormer \citep{li2023voxformer} & - & - & - & 38.76 & 11.91 \\
        OccFormer \citep{zhang2023occformer} & {31.39} & {19.03} & 21.93 & 40.27 & 13.81\\
        \midrule
        TPVFormer \citep{huang2023tpv} & {30.86} & 17.10 & 27.83 & - & - \\

        \rowcolor{Apricot!20!}
        \quad w/ \method{} & 32.05 & \textbf{20.85} & \textbf{30.48} &  - & - \\

        & \textcolor{ForestGreen}{+1.19} & \textcolor{ForestGreen}{+3.75} & \textcolor{ForestGreen}{+2.65} & - & - \\

        \midrule

        SurroundOcc \cite{wei2023surroundocc} & 31.49 & 20.30 & 29.21 & 38.51 & 13.08 \\

        \rowcolor{Apricot!20!}
        \quad w/ \method{} & \textbf{32.61} &  20.82 &  30.38 &  38.62 & 13.34 \\
        & \textcolor{ForestGreen}{+1.12} & \textcolor{ForestGreen}{+0.52} & \textcolor{ForestGreen}{+1.17} &  \textcolor{ForestGreen}{+0.11} & \textcolor{ForestGreen}{+0.26} \\

        \midrule
        Symphonies \citep{jiang2024symphonies} & - & - & - & 43.40 & 17.82 \\
        \rowcolor{Apricot!20!}
        \quad w/ \method{} & - & - & - & \textbf{44.08} & \textbf{18.11} \\
         & - & - & - & \textcolor{ForestGreen}{+0.68} & \textcolor{ForestGreen}{+0.29} \\
        \bottomrule
    \end{tabular}}
    \caption{\textbf{Performance Comparison on Multiple 3D Occupancy Benchmarks.} We report IoU ($\uparrow$) and mIoU ($\uparrow$) metrics on \textbf{SurroundOcc-nuScenes} \citep{wei2023surroundocc}, \textbf{Occ3D-nuScenes} \citep{tian2023occ3d}, and \textbf{SSCBench-KITTI360} \citep{li2024sscbench}. The best results are highlighted in bold. Our module, \method{}, consistently improves performance when integrated with standard models, achieving state-of-the-art results across all benchmarks. Performance gains introduced by \method{} are shown in green.}
    \label{tab:unified_benchmark}
\end{table}

%% file: tables/2D_training.tex
\begin{table}[t!]
    \centering
    \small
    \setlength{\tabcolsep}{5pt}
    \resizebox{\columnwidth}{!}{%
    \begin{tabular}{@{}lccccc@{}}
        \toprule
        \textbf{Methods} 
        & {Labels} 
        & \shortstack{{2D} \\ {GT Img.}} 
        & \shortstack{{Train Mem.} \\ {Overhead}} 
        & \shortstack{{Single} \\ {Frame}} 
        & {mIoU ($\uparrow$)} \\
        \midrule
        SelfOcc         & Pseudo-lbs. & \textcolor{green!60!black}{Dense}  & \textcolor{red}{High}  & \cmark & 9.3 \\
        OccNerf         & Pseudo-lbs. & \textcolor{green!60!black}{Dense}  & \textcolor{red}{High}  & \xmark & 9.5 \\
        GaussianOcc     & Pseudo-lbs. & \textcolor{green!60!black}{Dense}  & \textcolor{green!60!black}{Low}  & \cmark & 9.9 \\
        \midrule
        RenderOcc       & Lidar       & \textcolor{red}{Sparse}            & \textcolor{red}{High}  & \xmark & 19.3 (S) / 23.9 (T) \\
        \rowcolor{Apricot!20!}
        GaussRender     & Voxels      & \textcolor{green!60!black}{Dense}  & \textcolor{green!60!black}{Low}  & \cmark & \textbf{25.3} \\
        \bottomrule
    \end{tabular}
    }
    \caption{\textbf{Comparison of rendering-based methods on Occ3d-nuScenes under 2D-only supervision.} We report the mIoU on fully trained models. Labels include pseudo-labels, sparse Lidar, or dense voxel supervision. GaussRender achieves the best performance with low memory and using a dense image supervision signal.}
    \label{tab:rendering_methods_comparison}
\end{table}

%% file: tables/kitti360_additional.tex
\begin{table}[t!]
    \centering
    \small
    \setlength{\tabcolsep}{5pt}
    \renewcommand\theadfont{\normalsize\bfseries}
    \resizebox{\columnwidth}{!}{%
    \begin{tabular}{@{}lccccc@{}}
        \toprule
        & \makecell{3D all \\ mIoU / IoU} 
        & \makecell{3D visible \\ mIoU / IoU} 
        & \makecell{Image \\ mIoU / IoU} 
        & \makecell{Depth \\ L1} \\
        \midrule
        SurroundOcc \cite{wei2023surroundocc}
        & 13.08 / 38.51 
        & 25.83 / 69.40 
        & 34.20 / 90.72 
        & 1.394 \\
        
        \rowcolor{Apricot!20!} \hspace{.2cm} {w/ \method{}} 
        & 13.34 / 38.62 
        & 27.06 / 71.39 
        & 35.86 / 94.55 
        & 1.173 \\
        
        & \textcolor{green!60!black}{+0.26 / +0.11} 
        & \textbf{\textcolor{green!60!black}{+1.23 / +1.99}} 
        & \textbf{\textcolor{green!60!black}{+1.66 / +3.83}} 
        & \textcolor{green!60!black}{-0.221} \\
        \bottomrule
    \end{tabular}
    }
    \caption{\textbf{Performance improvement of SurroundOcc using GaussRender on KITTI-360.} GaussRender leads to clear gains even on challenging datasets, especially in visible 3D regions.}
    \label{tab:gaussrender_kitti360}
\end{table}

%% file: tables/rayiou.tex
\begin{table}[t]
    \centering
    \small
    \resizebox{0.9\columnwidth}{!}
    {
    \setlength{\tabcolsep}{5pt}
    \begin{tabular}{@{}lcc@{}}
    \toprule
    \textbf{Models} & RayIoU & RayIoU\textsubscript{1m, 2m, 4m} \\
    \midrule
    RenderOcc \cite{pan2024renderocc} & 19.5 & 13.4, 19.6, 25.5 \\
    BEVDet-Occ \cite{huang2022bevdet4d} & 29.6 & 23.6, 30.0, 35.1 \\
    BEVFormer \cite{li2022bevformer} & 32.4 & 26.1, 32.9, 38.0 \\
    BEVDet-Occ-Long \cite{huang2022bevdet4d} & 32.6 & 26.6, 33.1, 38.2 \\
    SparseOcc\cite{liu2024sparseocc} (8f) & 34.0 & 28.0, 34.7, 39.4 \\
    FB-Occ \cite{li2023fbocc3d} & 33.5 & 26.7, 34.1, 39.7 \\
    SparseOcc \cite{liu2024sparseocc} (16f) & {36.1} & {30.2, 36.8, 41.2} \\
    \hline
    TPVFormer \cite{huang2023tpv} & {37.2} & 31.5, 38.1, 41.9  \\
    \rowcolor{Apricot!20!} \hspace{.2cm} w/ GaussRender & \textbf{38.3} \textcolor{ForestGreen}{+1.1} & \textbf{32.3, 39.3, 43.4}  \\
    SurroundOcc \cite{wei2023surroundocc} & {35.5} & 29.9, 36.3, 40.1 \\
    \rowcolor{Apricot!20!} \hspace{.2cm} w/ GaussRender & \underline{37.5} \textcolor{ForestGreen}{+2.0} & \underline{31.4, 38.5, 42.6} \\
    \bottomrule
    \end{tabular}
    }
    \caption{\textbf{Impact of \method{} on RayIoU ($\uparrow$) metrics on the Occ3D-nuScenes validation dataset.} Best results are in bold. Previous results are reported from \cite{liu2024sparseocc}. $\text{RayIoU}$\textsubscript{1m, 2m, 4m} refers to the RayIoU with a depth tolerance of 1, 2, or 4 meters.}
    \label{tab:rayiou_metrics}
\end{table}

%% file: fig/camera_strategy.tex
\begin{table}[t!]
    \centering
    \small
    \setlength{\tabcolsep}{3pt}
    \resizebox{\columnwidth}{!}{%
    \begin{tabular}{@{}lccccc@{}}
        \hline
        \textbf{Cameras} & Sensor & Elevated & Elevated + Around. & Fully Rand. & Dynamic \\
        \hline
        & \includegraphics[width=0.18\linewidth]{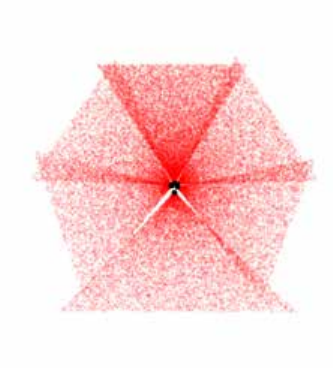} 
        & \includegraphics[width=0.18\linewidth]{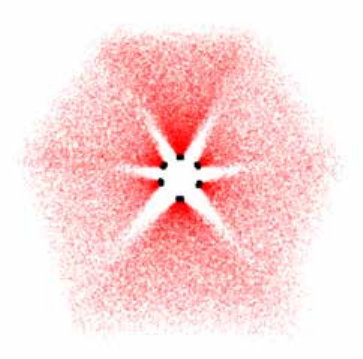} 
        & \includegraphics[width=0.18\linewidth]{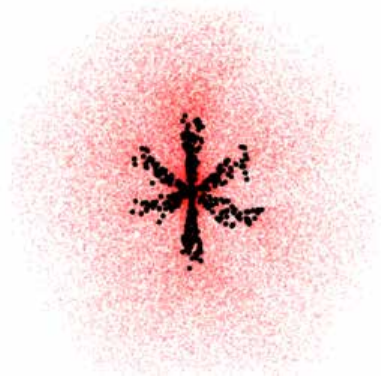} 
        & \includegraphics[width=0.18\linewidth]{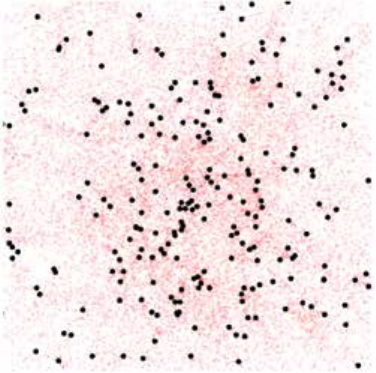} 
        & \includegraphics[width=0.18\linewidth]{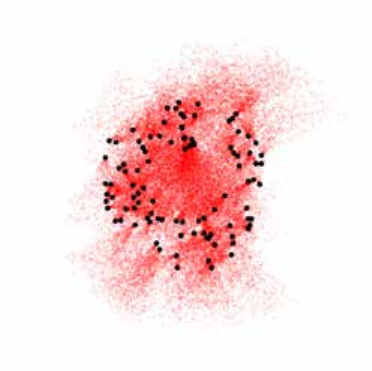} 
        \\
        2D+3D   & 25.9 & \underline{26.0} & \textbf{26.3} & 25.4 & 25.8 \\
        2D-only & \textbf{16.1} & 11.7 & \underline{12.9} & 6.9 & 4.8 \\
        \hline
    \end{tabular}}
    \caption{\textbf{Comparison of mIoU with different camera sampling strategies} trained and evaluated on 20\% of Occ3D. We evaluate five sampling strategies under 2D-only and 2D+3D supervision. Images illustrate the impact of camera positioning, black dots represent camera centers while red dots represent visible points casted in their frustum.}
    \label{tab:camera_sampling_strategies}
\end{table}

%% file: tables/loss_increment.tex
\begin{table}[t!]
    \centering
    \small
    \setlength{\tabcolsep}{3pt}
    \begin{tabular}{@{} cccc cc @{}}
        \hline
        \multicolumn{4}{c}{\fontsize{9}{9}\selectfont Loss Components} & \multicolumn{2}{c}{\fontsize{9}{9}\selectfont SurroundOcc-nuSc \citep{tian2023occ3d}} \\
        \cmidrule(lr){1-4} \cmidrule(lr){5-6}
        Cam & Depth & BEV & Bev Depth & IoU ($\uparrow$) & mIoU ($\uparrow$) \\
        \hline
        \checkmark & & & & 26.3 & 14.3 \\
        \checkmark & \checkmark & & & 26.8 & 15.1 \\
        \checkmark & \checkmark & \checkmark & & 27.2 & 15.6 \\
        \rowcolor{Apricot!20!} \checkmark & \checkmark & \checkmark & \checkmark & \textbf{27.5} & \textbf{16.4} \\
        \hline
    \end{tabular}
    \caption{\textbf{Impact of adding different loss components on 3D semantic occupancy performances.} The architecture used is TPVFormer \cite{huang2023tpv}. Models are trained with different combinations of losses and evaluated on 3D IoU and mIoU. We train the models on 20\% of SurroundOcc-nuScenes \cite{tian2023occ3d}.}
    \label{tab:loss_components_study}
\end{table}

%% file: sec/5_conclusion.tex
\section{Conclusion}
\label{sec:ccl}
{\method{} introduces a novel paradigm for enhancing 3D occupancy prediction through differentiable Gaussian rendering, bridging the gap between voxel-based supervision and projective consistency. By enforcing {projective} geometric and semantic coherence, 
it
significantly reduces spatial artifacts while maintaining compatibility with diverse architectures. The module's efficiency and flexibility enable state-of-the-art performance across benchmarks.}

{Looking ahead, several exciting directions emerge. First, integrating dynamic viewpoint synthesis with temporal sequences could further improve occlusion reasoning. Second, extending Gaussian rendering to open-vocabulary 3D understanding (leveraging vision-language models) might unlock semantic reasoning beyond predefined classes.}

\section{Acknowledgments}
This work was partially supported by the ANR MultiTrans project (ANR-21-CE23-0032). This work was granted access to the HPC resources of IDRIS under the allocation AD011014252R1 made by GENCI. We also thank Yihong Xu for his feedback on the paper and his valuable discussions.

%% file: sec/X_suppl.tex
\clearpage
\setcounter{page}{1}
\maketitlesupplementary

\appendix

\section{Datasets}
\label{app:datasets}

\paragraph{SurroundOcc-nuScenes \citep{wei2023surroundocc} and Occ3D-nuScenes \citep{tian2023occ3d}} are derived from the nuScenes dataset \cite{caesar2020nuscenes}. nuScenes provides 1000 scenes of surround view driving scenes in Boston and Singapore split in three sets train/val/test of size 700/150/150 scenes. Each comprises 20 seconds long and is fully annotated at 2Hz using one ground truth from 5 radars, 6 cameras at resolution $900 \times 1600$ pixels, one LiDAR, and one IMU.
From the LiDAR annotations, SurroundOcc-nuScenes \cite{wei2023surroundocc} derived a 3D grid of shape $[200,200,16]$ with a range in $[-50,50] \times [-50,50] \times [-5,3]$ meters at a spatial resolution of $[0.5,0.5,0.5]$ meters, annotated with 17 different classes, 1 representing empty and 16 for the semantics. 
The Occ3d-nuScenes \cite{tian2023occ3d} dataset has a lower voxel size of 0.4 meters in all directions while keeping the same voxel grid shape with a range of $[-40,40]\times [-40,40]\times [-1,5.4]$ meters. It contains 18 classes: 1 representing empty, 16 for the semantics, and 1 for others.
\paragraph{SSCBench-Kitti360 \cite{li2024sscbench}} is derived from the Kitti360 dataset \cite{Liao2022PAMI}. Kitti360 consists of over 320k images shot by 2 front cameras at a resolution $376 \times 1408$ pixels and two fisheye cameras in surburban areas covering a driving distance of 73.7km. Only one camera is used in the 3D occupancy task.
SSCBench-Kitti360 \cite{li2024sscbench} annotates for each sequence a voxel grid of shape $[256,256,32]$ with a range in $[0,51.2] \times [-25.6,25.6] \times [-2,4.4]$ meters at a voxel resolution of 0.2 in all directions.
{The provided voxel grid is annotated with 19 classes: one is used to designate empty voxels, and the 18 other are used for the semantic classes.}

\section{Models and implementation details}
\label{app:models}

We integrate our rendering module and associated loss into three different models: \textbf{SurroundOcc} \cite{wei2023surroundocc} (multi-scale voxel-based approach), \textbf{TPVFormer} \cite{huang2023tpv} (triplane-based approach), and \textbf{Symphonies} \cite{jiang2024symphonies} (voxel-with-instance query-based approach).
Each model is retrained using the same training setting, following the optimization parameters from SurroundOcc. No extensive hyperparameter searches are conducted on the learning rate; the goal is to demonstrate that the loss can be integrated at minimal cost into existing pipelines. All models are trained for 20 epochs on 4 A100 or H100 GPUs with a batch size of 1, using an AdamW optimizer with a learning rate of $2e^{-4}$ and a weight decay of $0.01$.
For each combination of models and datasets, we evaluate existing checkpoints if provided; otherwise, we report the scores from previous papers when available or we re-train the models. Note that we used the official checkpoint for Symphonies \cite{jiang2024symphonies} while noticing there is a discrepancy in IoU / mIoU between the reported value in the paper and the actual one of the official checkpoint, as explained in their GitHub issue \footnote{https://github.com/hustvl/Symphonies/issues/5}.

\section{Computational cost}

{Our module introduces a computation overhead for each rendering it performs. For a given input scene, we generate two views (one `cam' and one `bev'), and for each view, we render both the predictions and the ground truth, resulting in a total of four renderings per iteration.
}

{
As reported in \cref{tab:gpu_memory_comparison}, training with \method{} incurs a modest increase in memory and computation time (~10\%), while using high-resolution renderings. This overhead can be further reduced by pre-selecting camera locations, allowing annotation renderings (the two ground-truth renderings) to be pre-processed in advance. Additionally, lower rendering resolutions can be used if needed.
}

{While \method{} introduces a small per-iteration cost, it actually accelerates learning. A key observation is that models using \method{} reach the same performance level as their baseline counterpart ~17\% faster. Overall, despite a minor increase in computational overhead, \method{} ultimately reduces the total training time required to achieve comparable or superior performance.}

\input{tables/computational_cost}

\section{BeV metrics.}
\input{tables/image_depth_bev_metrics}

Additionally, we evaluate some Bird’s-Eye-View (BeV) metrics --- critical for {downstream} motion forecasting and planning --- measuring spatial accuracy {on the horizontal plane} using IoU$^\text{BeV}$ and mIoU$^\text{BeV}$ that capture different aspects of spatial understanding.

{
For this study, we compute the orthographic BeV image for both the prediction and the ground truth, following the rendering procedure of \autoref{sec:methods:gaussian_rendering} for ground truth.
The class assigned to the pixel $p$ is the one corresponding to the maximal value of $C_p$.
Then, we compute IoU (binary occupancy, full vs empty) and mIoU (semantic occupancy) by comparing the two images.
}

Our analysis is presented in \autoref{tab:bev_metrics}.
We observe that the use of \method{} enhances {both} metrics simultaneously, with systematic gains associated with different combinations of datasets and models. 
Both TPVFormer and SurroundOcc show significant improvements across all datasets: Occ3d-nuScenes and SurroundOcc-nuScenes and evaluations.
This evaluation highlights that the use of \method{} not only improves 3D occupancy predictions but also enhances consistency with BeV and sensor observations.

\section{Ablations}
\subsection{Gaussian scaling}

An important parameter in our rendering process is the fixed size of the Gaussians representing voxels. To study its impact, we train a TPVFormer \cite{huang2023tpv} model on Occ3d-nuScenes \cite{tian2023occ3d}, varying the Gaussian scale for both ground-truth and predicted renderings. We train models using only the 2D rendering losses (\autoref{eq:rendering_losses}), excluding the usual 3D voxel losses to isolate the effect of scale on rendering metrics.

\input{fig/impact_of_scale_fig}

Our results, shown in \autoref{fig:impact_of_scale}, highlight the importance of the Gaussian scale. If the Gaussians are too large, only a few will cover the image, and the loss will be backpropagated mainly from the nearest ones. If they are too small, gaps appear between voxels, leading to sparse activations and a model that renders mostly the empty class, yielding poor metrics.

Theoretically, the optimal size should correlate with the voxel size. For Occ3d-nuScenes and SurroundOcc-nuScenes, the optimal scale is $s = 0.25$ and $s = 0.20$, while for SSCBench-KITTI360, it is $s = 0.1$. This aligns with our intuition: a voxel should be represented by a spherical Gaussian with a standard deviation such that $2s = c$, where $c$ is the voxel side.

Qualitatively, \autoref{fig:impact_of_scale} shows the effect of scale on rendering, confirming the need for a balanced Gaussian size to avoid either sparse activation or excessive concentration on nearby elements.

\input{fig/visu_scale_render}

\subsection{Loss balance}

We {investigate the importance of the balance} $\lambda$ between the 2D loss and the 3D loss. If the {weight of the} 2D loss is too high, there is a risk of optimizing the image rendering at the expense of voxel predictions. Conversely, if the 2D loss is too low, its contribution to the learning process may be {overlooked}. To analyze this, we vary the {contribution $\lambda$ of the 2D loss} and study the impact on the final metrics, {as reported in} \autoref{fig:lambda_impact}. Based on the training results of TPVFormer \cite{huang2023tpv} on a subset of Occ3D-nuScenes \cite{tian2023occ3d}, we set the weight to $\lambda=15$.
\input{tables/lambda_loss}

\subsection{Cameras}
\label{app:ablation:camera}

{
For a given input, we can position as many cameras as needed to render multiple views. In this experiment, we explore using multiple cameras by selecting from the six available views.
While, in theory, more cameras could provide more accurate gradients, we observe in practice that it does not significantly impact the final results (\autoref{fig:num_cams_impact}).
Since additional cameras introduce computational overhead, we opt to render from a single camera per iteration, changing its position across batches according to the strategy defined in \autoref{sec:methods:camera}.
}

\subsection{Camera strategies}
\label{app:camera_strat}

\begin{itemize}
\item \textbf{Sensor Strategy:} Cameras are placed at the original sensor locations and orientations as provided in the dataset.
\item \textbf{Elevated Strategy:} Each camera is lifted and tilted downward. This modification increases the vertical field of view, providing a top-down perspective that reduces self-occlusion and captures a broader context of the scene.
\item \textbf{Elevated + Around Strategy:} This strategy combines elevation and downward tilt with additional random displacements around the ego vehicle—up to half the maximum scene range. It allows observing the scene from novel angles while maintaining a consistent overhead viewpoint, improving the visibility of occluded voxels.
\item \textbf{Fully Random Strategy:} Cameras are randomly placed throughout the scene, applying pitch and yaw perturbations and varying distances from the ego-vehicle. While this increases the diversity of viewpoints, it also introduces inconsistency and often places cameras in less informative positions (e.g., viewing empty space).
\item \textbf{Dynamic Strategy:} Cameras are elevated and positioned at random distances along a circular ring. However, each camera is oriented to look at the element with the highest 3D cross-entropy. In other words, the camera focuses on a region where it made the largest prediction error. It appears that its supervision signal does not help much the training.
\end{itemize}

\input{tables/number_of_cameras}

\section{Scores detailed per class}
\label{app:results}

The following tables give the detailed IoU and mIoU scores of the models studied for each dataset. \autoref{tab:Occ3d-nuScenes} concerns the Occ3D-nuScenes dataset \cite{tian2023occ3d}, \autoref{tab:SSCBench-KITTI360} the SSCBench-KITTI360 dataset \cite{li2024sscbench} and  \autoref{tab:SurroundOcc-Nuscenes} the SurroundOcc-nuScenes dataset \cite{wei2023surroundocc}. The variations by class appear to be due to learning variance, which is why it makes more sense to look at the overall IoU and RayIoU metrics, rather than looking for intrinsic reasons.

\section{Qualitative results}
\label{app:qualitative}
In \autoref{fig:qualitative_0} and \autoref{fig:qualitative_1}, we respectively present qualitative results on randomly selected scenes from SurroundOcc-nuScenes \cite{wei2023surroundocc} and Occ3d-nuScenes datasets \cite{tian2023occ3d}. We also provide complete gifs in our github.

\input{tables/surr_results}
\input{tables/occ3D_results}
\input{tables/kitti360_results}

\include{fig/qualitative}

%% file: tables/computational_cost.tex
\begin{table}[!t]
\centering
\small
    {%

    \setlength{\tabcolsep}{5pt}

\begin{tabular}{@{}c l cccc}
    \toprule
    \multirow{2}{*}{\rotatebox[origin=c]{90}{\fontsize{8}{8}\selectfont Data}} & & \multicolumn{2}{c}{Tr. time} & \multicolumn{2}{c}{Memory usage} \\
          &  Model    & \multicolumn{2}{c}{(HH:MM)}  & \multicolumn{2}{c}{(GB)} \\
    \midrule
    & TPVFormer  & 21:44 &         & 25.3GB & \\
    \rowcolor{Apricot!20!}\cellcolor{white}
    & ~~~~ w/ \method{}    & 24:00 & +10.4\% & 28.1GB & +11.1\% \\
    & SurroundOcc& 26:38 &         & 23.0GB & \\
    \rowcolor{Apricot!20!}\cellcolor{white}
    \multirow{-4}{*}{\rotatebox[origin=c]{90}{\fontsize{8}{8}\selectfont Sur.Occ-nusc}} & ~~~~ w/ \method{}     & 29:19 & +10.5\% & 24.2GB & +5.2\% \\ 
    \midrule
    & TPVFormer   & 7:02 & & 29.3GB \\
    \rowcolor{Apricot!20!}\cellcolor{white}
    & ~~~~ w/ \method{}    & 8:16 & +14.0\%  & 31.7GB & +8.2\% \\
    & SurroundOcc & 11:12  & & 15.5GB \\
    \rowcolor{Apricot!20!}\cellcolor{white}
    \multirow{-4}{*}{\rotatebox[origin=c]{90}{\fontsize{8}{8}\selectfont SSCB.K.360}}
    &  ~~~~ w/ \method{}     & 11:56 & +6.1\% & 17.6GB & +13.5\% \\
    \bottomrule
\end{tabular}
}
\caption{\textbf{Training time and GPU memory usage} across models and datasets without or with our module using four renderings per scene, two for BeV (ground truth and predictions), and two for another camera (ground truth and predictions). Test performed on a 40GB A100.}
\label{tab:gpu_memory_comparison}
\end{table}

%% file: tables/image_depth_bev_metrics.tex
\begin{table}[t]
    \centering
    \small
{
    \setlength{\tabcolsep}{3pt}
    \begin{tabular}{@{}l l l l@{}}
        \toprule
        Dataset & Model & IoU$^\text{BeV}$ ($\uparrow$) & mIoU$^\text{BeV}$ ($\uparrow$) \\
        \midrule
        
        & TPVFormer \citep{huang2023tpv} & 58.16 & 28.48 \\
        \rowcolor{Apricot!20!}\cellcolor{white}SurroundOcc- 
        & \hspace{0.3cm} w/ \method{} & 59.20 \textcolor{ForestGreen}{+1.04} & \textbf{28.73} \textcolor{ForestGreen}{+0.25} \\

        nuSc \citep{wei2023surroundocc} & SurroundOcc \citep{wei2023surroundocc} & 58.60 & 28.26 \\
        \rowcolor{Apricot!20!} \cellcolor{white}
         & \hspace{0.3cm} w/ \method{} & \textbf{60.55} \textcolor{ForestGreen}{+1.95} & 28.64 \textcolor{ForestGreen}{+0.38}\\
        \midrule
        
        & TPVFormer \citep{huang2023tpv} & 52.95 & 29.72 \\
        
        \rowcolor{Apricot!20!}  \cellcolor{white}Occ3D- 
        & \hspace{0.3cm} w/ \method{} & 54.35 \textcolor{ForestGreen}{+1.40} & 30.26 \textcolor{ForestGreen}{+0.54} \\
        
        nusc \citep{tian2023occ3d} & SurroundOcc \citep{wei2023surroundocc} & 53.52 & 28.98 \\
        \rowcolor{Apricot!20!} \cellcolor{white}
         &  \hspace{0.3cm} w/ \method{} & \textbf{55.65} \textcolor{ForestGreen}{+2.13} & \textbf{30.50} \textcolor{ForestGreen}{+1.52} \\
        \bottomrule
    \end{tabular}
}
    
    \caption{\textbf{Impact of \method{} on BeV metrics.} Comparison of BeV metrics (IoU$^\text{BeV}$, mIoU$^\text{BeV}$) across datasets. Best results per dataset/metric are in bold with green performance deltas. %
    }
    \label{tab:bev_metrics}
\end{table}

%% file: fig/impact_of_scale_fig.tex
\begin{figure}[!t]
    \centering
\begin{tikzpicture}
    \begin{axis}[
        width=\columnwidth, %
        height=5cm, %
        xlabel={Gaussian scale ($s$)},
        ylabel={Value (\%)},
        xmin=0, xmax=0.51,
        ymin=0, ymax=45,
        xtick={0.01, 0.10, 0.20, 0.25, 0.30, 0.40, 0.50},
        xticklabels={0.01, 0.10, 0.20, 0.25, 0.30, 0.40, 0.50},
        ytick={0, 10, 20, 30, 40}, %
        legend style={at={(0.7, 0.3)}}, %
        grid=both,
        grid style={line width=.1pt, draw=gray!10},
        major grid style={line width=.2pt, draw=gray!50},
    ]
        \addplot[
            color=cyan,
            mark=*,
            mark options={solid, fill=cyan},
        ] coordinates {
            (0.01, 0.0)
            (0.10, 37.8)
            (0.20, 40.2)
            (0.25, 41.3)
            (0.30, 42.8)
            (0.40, 42.7)
            (0.50, 0.0)
        };
        \addlegendentry{IoU ($\uparrow$)}

        \addplot[
            color=red,
            mark=square*,
            mark options={solid, fill=red},
        ] coordinates {
            (0.01, 0.0)
            (0.10, 12.9)
            (0.20, 15.7)
            (0.25, 15.6)
            (0.30, 15.9)
            (0.40, 15.0)
            (0.50, 0.0)
        };
        \addlegendentry{mIoU ($\uparrow$)}
    \end{axis}
\end{tikzpicture}
    \caption{\textbf{Impact of fixed Gaussian scales on 3D mIoU and IoU} using TPVFormer \cite{huang2023tpv} trained using only $L_{\text{2D}}$ without $L_{\text{3D}}$ on 20\% of Occ3d-nuScenes \cite{tian2023occ3d} validation dataset.}
    \label{fig:impact_of_scale}
\end{figure}
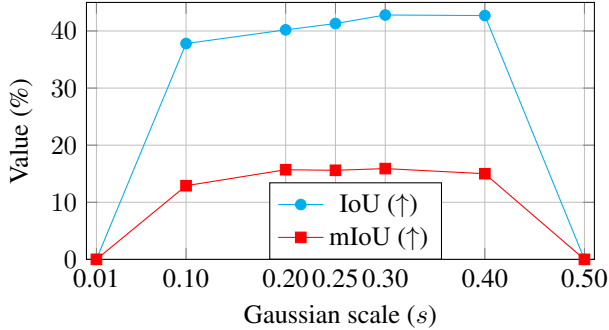

%% file: fig/visu_scale_render.tex
\begin{figure}[!t]
    \centering
    \begin{subfigure}{0.32\linewidth}
        \includegraphics[width=\linewidth]{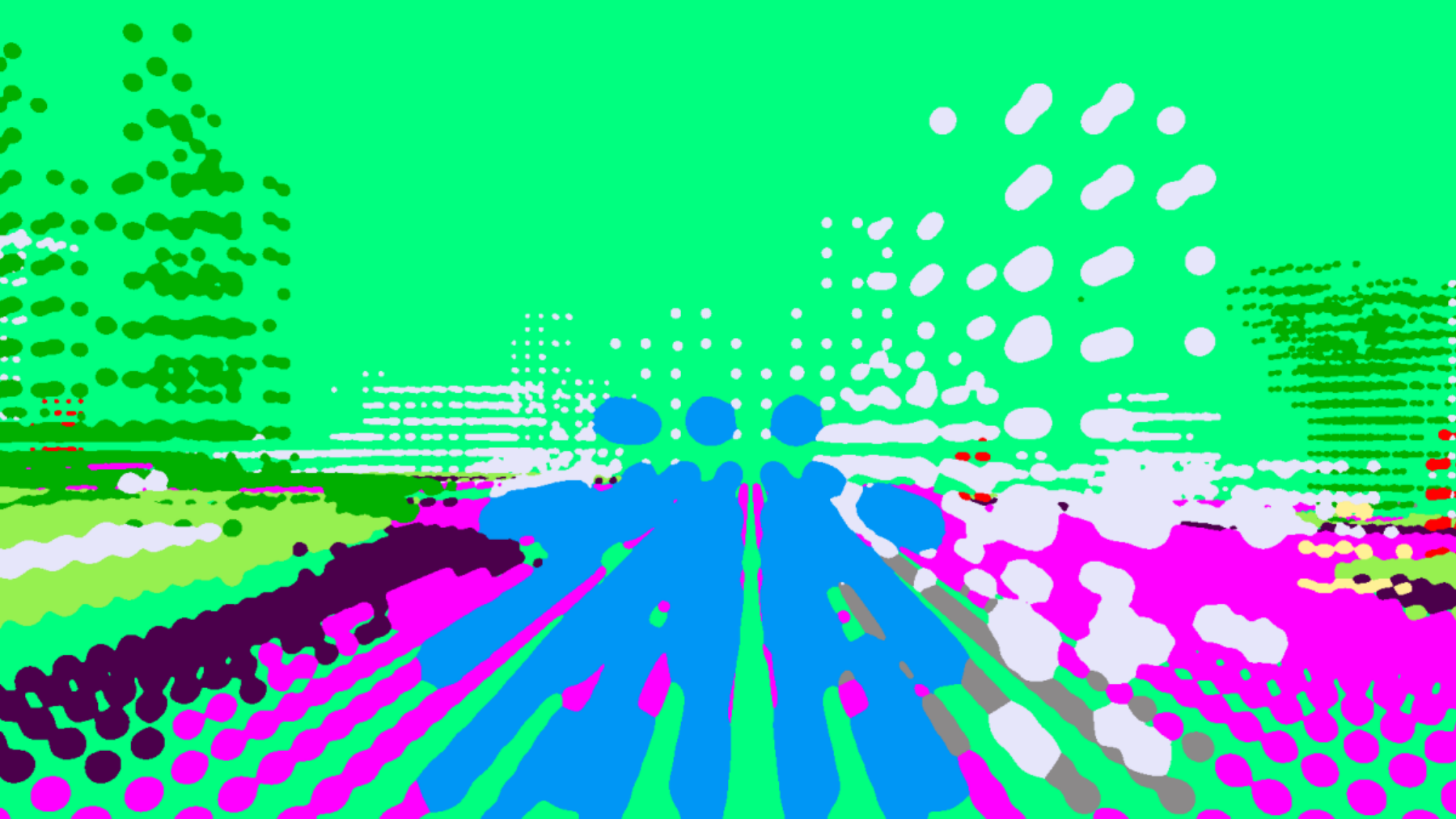}
        \caption{$\sigma = 0.06$}
    \end{subfigure}
    \hfill
    \begin{subfigure}{0.32\linewidth}
        \includegraphics[width=\linewidth]{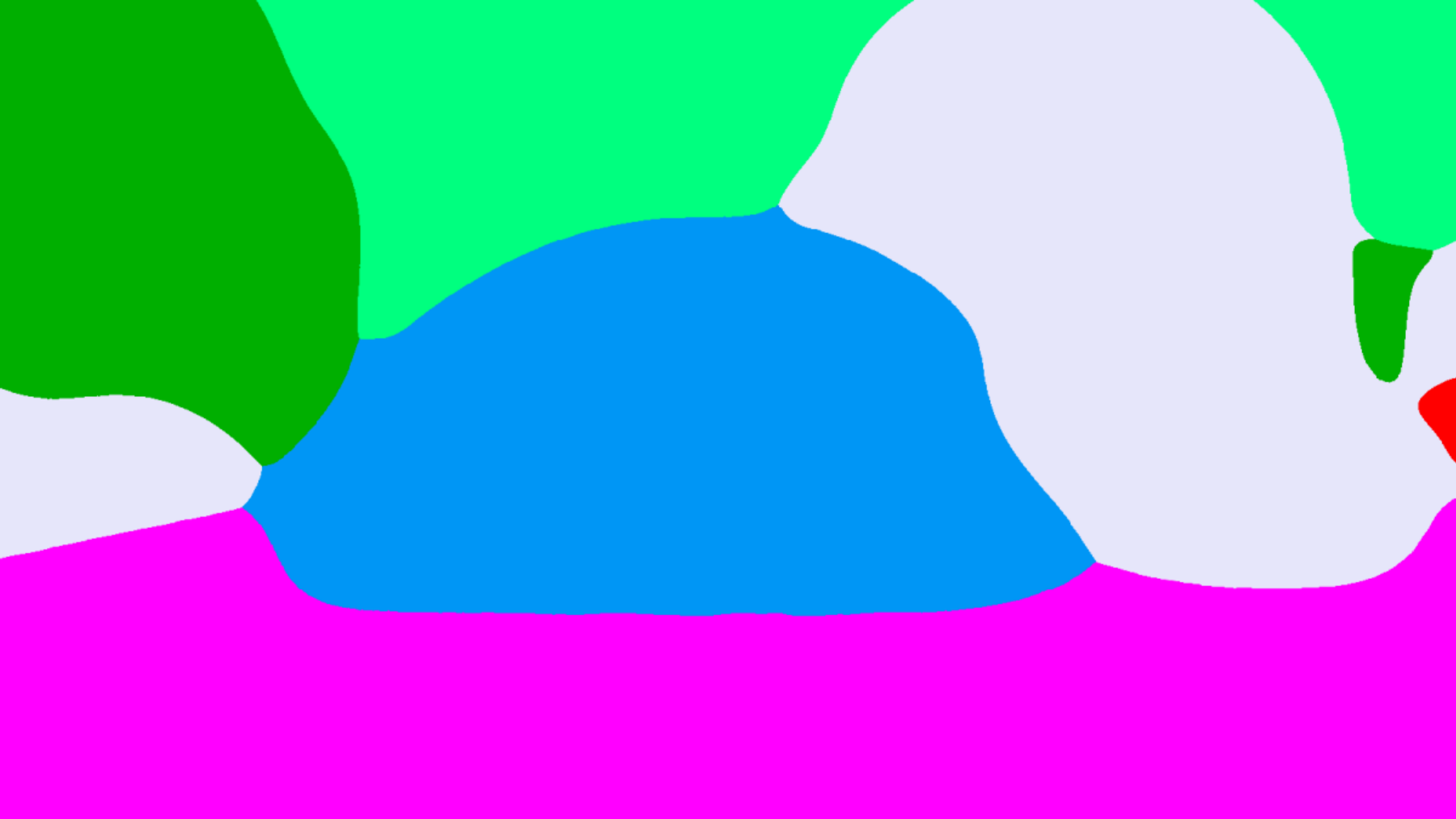}
        \caption{$\sigma = 0.2$}
    \end{subfigure}
    \hfill
    \begin{subfigure}{0.32\linewidth}
        \includegraphics[width=\linewidth]{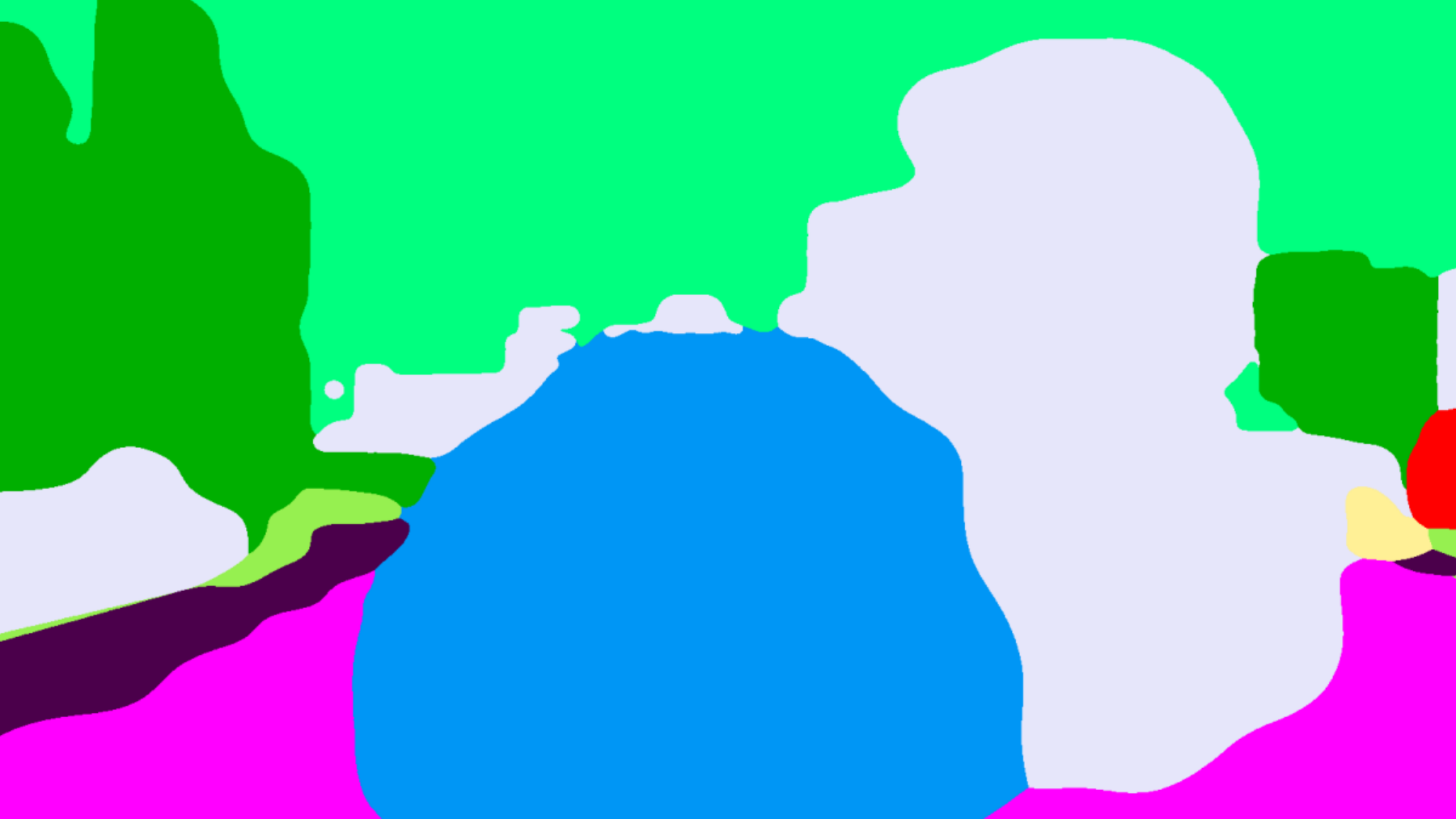}
        \caption{$\sigma = 0.4$}
    \end{subfigure}

    \vspace{1em} %

    \begin{subfigure}{0.48\linewidth}
        \includegraphics[width=\linewidth]{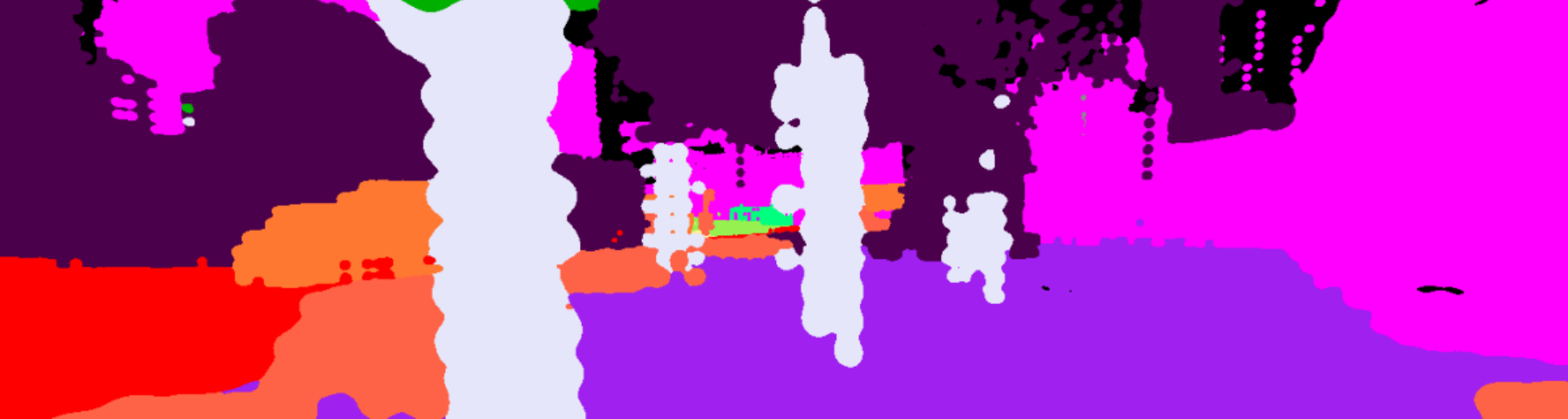}
        \caption{$\sigma = 0.06$}
    \end{subfigure}
    \hfill
    \begin{subfigure}{0.48\linewidth}
        \includegraphics[width=\linewidth]{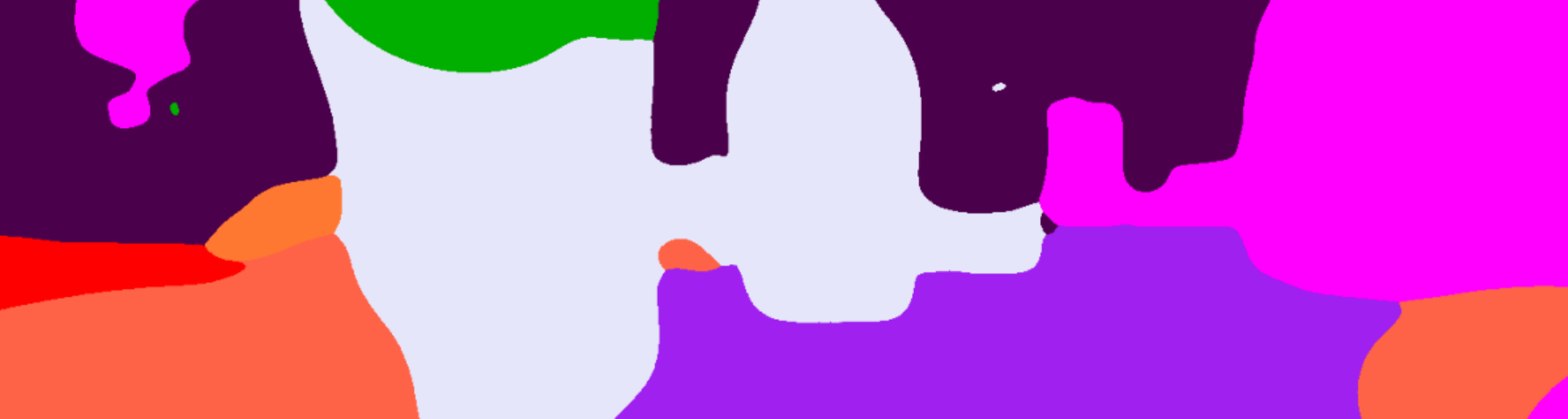}
        \caption{$\sigma = 0.2$}
    \end{subfigure}

    \caption{\textbf{Visualization of different Gaussianized voxels for different datasets and scales}. The first row represents data from Occ3d-nuScenes \cite{tian2023occ3d} and the second and third rows are from SSCBench-Kitti360 \cite{li2024sscbench}.}
    \label{fig:scale_render}
\end{figure}

%% file: tables/lambda_loss.tex
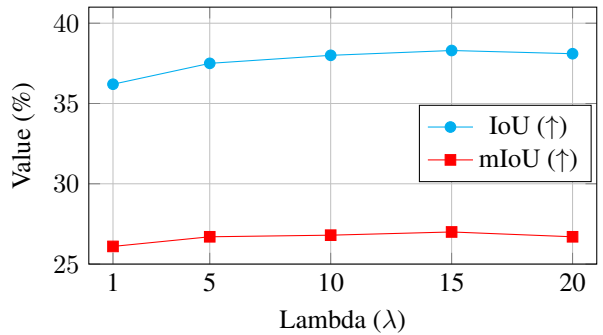
\begin{figure}[!t]
    \centering
\begin{tikzpicture}
    \begin{axis}[
        width=\columnwidth,
        height=5cm,
        xlabel={Lambda ($\lambda$)},
        ylabel={Value (\%)},
        xmin=0, xmax=21,
        ymin=25, ymax=41,
        xtick={1, 5, 10, 15, 20},
        xticklabels={1, 5, 10, 15, 20},
        ytick={25, 30, 35, 40},
        legend style={at={(0.99, 0.62)}},
        grid=both,
        grid style={line width=.1pt, draw=gray!10},
        major grid style={line width=.2pt, draw=gray!50},
    ]
        \addplot[
            color=cyan,
            mark=* ,
            mark options={solid, fill=cyan},
        ] coordinates {
            (1, 36.2)
            (5, 37.5)
            (10, 38.0)
            (15, 38.3)
            (20, 38.1)
        };
        \addlegendentry{IoU ($\uparrow$)}

        \addplot[
            color=red,
            mark=square*,
            mark options={solid, fill=red},
        ] coordinates {
            (1, 26.1)
            (5, 26.7)
            (10, 26.8)
            (15, 27.0)
            (20, 26.7)
        };
        \addlegendentry{mIoU ($\uparrow$)}
    \end{axis}
\end{tikzpicture}
    \caption{\textbf{Impact of different the contribution $\lambda$ of $L_{2D}$ on 3D semantic occupancy performance.} The architecture used is TPVFormer \cite{huang2023tpv}. Models are trained using a combination of $L_{\text{2D}}$ and $L_{\text{3D}}$ with varying $\lambda$ values and evaluated on 3D IoU and mIoU. We train and evaluate the models on 20\% of Occ3D-nuScenes \cite{tian2023occ3d} training and validation datasets.}
    \label{fig:lambda_impact}
\end{figure}

%% file: tables/number_of_cameras.tex
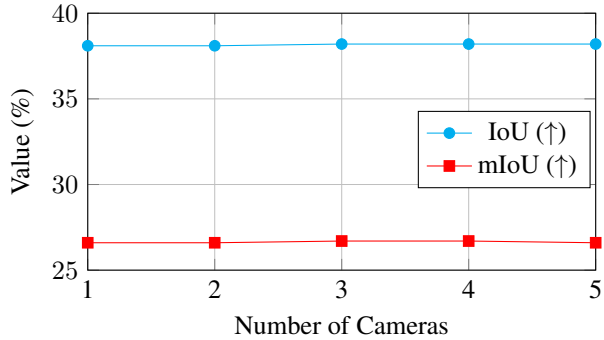
\begin{figure}[!t]
    \centering
\begin{tikzpicture}
    \begin{axis}[
        width=\columnwidth,
        height=5cm,
        xlabel={Number of Cameras},
        ylabel={Value (\%)},
        xmin=1, xmax=5,
        ymin=25, ymax=40,
        xtick={1, 2, 3, 4, 5},
        xticklabels={1, 2, 3, 4, 5},
        ytick={25,30,35, 40},
        legend style={at={(0.99, 0.62)}},
        grid=both,
        grid style={line width=.1pt, draw=gray!10},
        major grid style={line width=.2pt, draw=gray!50},
    ]
        \addplot[
            color=cyan,
            mark=* ,
            mark options={solid, fill=cyan},
        ] coordinates {
            (1, 38.1)
            (2, 38.1)
            (3, 38.2)
            (4, 38.2)
            (5, 38.2)
        };
        \addlegendentry{IoU ($\uparrow$)}

        \addplot[
            color=red,
            mark=square*,
            mark options={solid, fill=red},
        ] coordinates {
            (1, 26.6)
            (2, 26.6)
            (3, 26.7)
            (4, 26.7)
            (5, 26.6)
        };
        \addlegendentry{mIoU ($\uparrow$)}
    \end{axis}
\end{tikzpicture}
    \caption{\textbf{Impact of the number of cameras on 3D semantic occupancy performance.} The architecture used is TPVFormer \cite{huang2023tpv}. Models are trained with varying numbers of cameras and evaluated on 3D IoU and mIoU using 20\% of Occ3d-nuScenes training and validation set.}
    \label{fig:num_cams_impact}
\end{figure}

%% file: tables/surr_results.tex
\begin{table*}[t] %
    \small
    \setlength{\tabcolsep}{0.005\linewidth}  
    \centering
    \resizebox{\textwidth}{!}{
    \begin{tabular}{l|c c | c c c c c c c c c c c c c c c c}
        \toprule
        Model
 &  IoU
 &  mIoU
 &  \rotatebox{90}{\textcolor{nbarrier}{$\blacksquare$} barrier}
 &  \rotatebox{90}{\textcolor{nbicycle}{$\blacksquare$} bicycle}
 &  \rotatebox{90}{\textcolor{nbus}{$\blacksquare$} bus}
 &  \rotatebox{90}{\textcolor{ncar}{$\blacksquare$} car}
 &  \rotatebox{90}{\textcolor{nconstruct}{$\blacksquare$} const. veh.}
 &  \rotatebox{90}{\textcolor{nmotor}{$\blacksquare$} motorcycle}
 &  \rotatebox{90}{\textcolor{npedestrian}{$\blacksquare$} pedestrian}
 &  \rotatebox{90}{\textcolor{ntraffic}{$\blacksquare$} traffic cone}
 &  \rotatebox{90}{\textcolor{ntrailer}{$\blacksquare$} trailer}
 &  \rotatebox{90}{\textcolor{ntruck}{$\blacksquare$} truck}
 &  \rotatebox{90}{\textcolor{ndriveable}{$\blacksquare$} drive. suf.}
 &  \rotatebox{90}{\textcolor{nother}{$\blacksquare$} other flat}
 &  \rotatebox{90}{\textcolor{nsidewalk}{$\blacksquare$} sidewalk}
 &  \rotatebox{90}{\textcolor{nterrain}{$\blacksquare$} terrain}
 &  \rotatebox{90}{\textcolor{nmanmade}{$\blacksquare$} manmade}
 &  \rotatebox{90}{\textcolor{nvegetation}{$\blacksquare$} vegetation}
        \\
        \midrule
        MonoScene \cite{cao2022monoscene} & 23.96 & 7.31 & 4.03 &	0.35& 8.00& 8.04&	2.90& 0.28& 1.16&	0.67&	4.01& 4.35&	27.72&	5.20& 15.13&	11.29&	9.03&	14.86 \\
        Atlas \cite{murez2020atlas} & 28.66 & 15.00 & 10.64&	5.68&	19.66& 24.94& 8.90&	8.84&	6.47& 3.28&	10.42&	16.21&	34.86&	15.46&	21.89&	20.95&	11.21&	20.54 \\
        BEVFormer \cite{li2022bevformer} & 30.50 & 16.75 & 14.22 &	6.58 & 23.46 & 28.28& 8.66 &10.77& 6.64& 4.05& 11.20&	17.78 & 37.28 & 18.00 & 22.88 & 22.17 & {13.80} &	{22.21}\\
        TPVFormer-lidar \citep{huang2023tpv} & 11.51 & 11.66 & 16.14&	7.17& 22.63	& 17.13 & 8.83 & 11.39 & 10.46 & 8.23&	9.43 & 17.02 & 8.07 & 13.64 & 13.85 & 10.34 & 4.90 & 7.37\\
        OccFormer \citep{zhang2023occformer} & {31.39} & {19.03} & {18.65} & {10.41} & {23.92} & {30.29} & {10.31} & {14.19} & {13.59} & {10.13} & {12.49} & {20.77} & {38.78} & 19.79 & 24.19 & 22.21 & {13.48} & {21.35}\\
        GaussianFormer \citep{huang2024gaussian} & 29.83 & {19.10} & {19.52} & {11.26} & {26.11} & {29.78} & {10.47} & {13.83} & {12.58} & {8.67} & {12.74} & {21.57} & {39.63} & {23.28} & {24.46} & {22.99} & 9.59 & 19.12 \\
        GaussianFormerv2 \citep{huang2024gaussianformer2} & 30.56 & 20.02 & 20.15 & 12.99 & 27.61 & 30.23 & {11.19} & 15.31 & 12.64 & 9.63 & 13.31 & 22.26 & 39.68 & 23.47 & 25.62 & 23.20 & 12.25 & 20.73 \\ 
        \midrule
        TPVFormer \citep{huang2023tpv} & {30.86} & 17.10 & 15.96&	 5.31& 23.86	& 27.32 & 9.79 & 8.74 & 7.09 & 5.20& 10.97 & 19.22 & {38.87} & {21.25} & {24.26} & {23.15} & 11.73 & 20.81\\
        
        \rowcolor{Apricot!20!}
        \hspace{0.3cm} w/ \method{} & 32.05 & \textbf{20.85} & 20.2 & 13.06 & \textbf{28.95} & 30.96 & \textbf{11.26} & \textbf{16.69} & 13.64 & 10.57 & 12.77 & 22.58 & 40.69 & 23.49 & \textbf{26.41} & 24.97 & 14.41 & 22.94 \\        
        
        (gain) & \textcolor{ForestGreen}{1.19} & \textcolor{ForestGreen}{3.75} & \textcolor{ForestGreen}{4.24} & \textcolor{ForestGreen}{7.75} & \textcolor{ForestGreen}{5.09} & \textcolor{ForestGreen}{3.64} & \textcolor{ForestGreen}{1.47} & \textcolor{ForestGreen}{7.95} & \textcolor{ForestGreen}{6.55} & \textcolor{ForestGreen}{5.37} & \textcolor{ForestGreen}{1.80} & \textcolor{ForestGreen}{3.36} & \textcolor{ForestGreen}{1.82} & \textcolor{ForestGreen}{2.24} & \textcolor{ForestGreen}{2.15} & \textcolor{ForestGreen}{1.82} & \textcolor{ForestGreen}{2.68} & \textcolor{ForestGreen}{2.13} \\ \midrule

        SurroundOcc \cite{wei2023surroundocc} & 31.49 & 20.30 & 20.59 & 11.68 & 28.06 & 30.86 & 10.70 & 15.14 & \textbf{14.09} & \textbf{12.06} & \textbf{14.38} & 22.26 & 37.29 & 23.70 & 24.49 & 22.77 & 14.89 & 21.86 \\
        
        \rowcolor{Apricot!20!}
        \hspace{0.3cm} w/ \method{} & \textbf{32.61} & 20.82 & \textbf{20.32} &\textbf{ 13.22} & 28.32 & \textbf{31.05} & 10.92 & 15.65 & 12.84 & 8.91 & 13.29 & \textbf{22.76} & \textbf{41.22} & \textbf{24.48} & 26.38 & \textbf{25.20} & \textbf{15.31} & \textbf{23.25} \\
        
        (gain) & \textcolor{ForestGreen}{1.12} & \textcolor{ForestGreen}{0.52} & \textcolor{red}{-0.27} & \textcolor{ForestGreen}{1.54} & \textcolor{ForestGreen}{0.26} & \textcolor{ForestGreen}{0.19} & \textcolor{ForestGreen}{0.22} & \textcolor{ForestGreen}{0.51} & \textcolor{red}{-1.25} & \textcolor{red}{-3.15} & \textcolor{red}{-1.09} & \textcolor{ForestGreen}{0.50} & \textcolor{ForestGreen}{3.93} & \textcolor{ForestGreen}{0.78} & \textcolor{ForestGreen}{1.89} & \textcolor{ForestGreen}{2.43} & \textcolor{ForestGreen}{0.42} & \textcolor{ForestGreen}{1.39} \\
        \bottomrule
    \end{tabular}}
    \caption{\textbf{Semantic voxel occupancy results on the \textbf{SurroundOcc-NuScenes} \citep{wei2023surroundocc} validation set.} The best results are in bold. Training models with our module \method{} achieves state-of-the-art performance. Previous results are reported from \cite{huang2024gaussianformer2}.}
    \label{tab:SurroundOcc-Nuscenes}
\end{table*}

%% file: tables/occ3D_results.tex
\begin{table*}[t] %
    \small
    \setlength{\tabcolsep}{0.005\linewidth}  
    \centering
    \resizebox{\textwidth}{!}{
    \begin{tabular}{l|c|c|c|cccccccccccccccc}
        \hline
        Method & Input & mIoU 
 &  \rotatebox{90}{\textcolor{nother}{$\blacksquare$} others }
 &  \rotatebox{90}{\textcolor{nbarrier}{$\blacksquare$} barrier}
 &  \rotatebox{90}{\textcolor{nbicycle}{$\blacksquare$} bicycle}
 &  \rotatebox{90}{\textcolor{nbus}{$\blacksquare$} bus}
 &  \rotatebox{90}{\textcolor{ncar}{$\blacksquare$} car}
 &  \rotatebox{90}{\textcolor{nconstruct}{$\blacksquare$} const. veh.}
 &  \rotatebox{90}{\textcolor{nmotor}{$\blacksquare$} motorcycle}
 &  \rotatebox{90}{\textcolor{npedestrian}{$\blacksquare$} pedestrian}
 &  \rotatebox{90}{\textcolor{ntraffic}{$\blacksquare$} traffic cone}
 &  \rotatebox{90}{\textcolor{ntrailer}{$\blacksquare$} trailer}
 &  \rotatebox{90}{\textcolor{ntruck}{$\blacksquare$} truck}
 &  \rotatebox{90}{\textcolor{ndriveable}{$\blacksquare$} drive. suf.}
 &  \rotatebox{90}{\textcolor{nother}{$\blacksquare$} other flat}
 &  \rotatebox{90}{\textcolor{nsidewalk}{$\blacksquare$} sidewalk}
 &  \rotatebox{90}{\textcolor{nterrain}{$\blacksquare$} terrain}
 &  \rotatebox{90}{\textcolor{nmanmade}{$\blacksquare$} manmade}
 &  \rotatebox{90}{\textcolor{nvegetation}{$\blacksquare$} vegetation} \\ \hline
        MonoScene & Voxels & 6.06 & 1.75 & 7.23 & 4.26 & 4.93 & 9.38 & 5.67 & 3.98 & 3.01 & 5.90 & 4.45 & 7.17 & 14.91 & 6.32 & 7.92 & 7.43 & 1.01 & 7.65 \\
        BEVDet & Voxels & 19.38 & 4.39 & 30.31 & 0.23 & 32.26 & 34.47 & 12.97 & 10.34 & 10.36 & 6.26 & 8.93 & 23.65 & 52.27 & 24.61 & 26.06 & 22.31 & 15.04 & 15.10 \\
        OccFormer & Voxels  & 21.93 & 5.94& 30.29 & 12.32 & 34.40 & 39.17 & 14.44 & 16.45 & 17.22 & 9.27 & 13.90 & 26.36 & 50.99 & 30.96 & 34.66 & 22.73 & 6.76 & 6.97  \\
        BEVStereo& Voxels & 24.51 & 5.73& 38.41 &7.88 & 38.70 & 41.20 &17.56 &17.33 &14.69 &10.31 &16.84 &29.62 &54.08 &28.92 &32.68 &26.54 &18.74 &17.49 \\ 
        BEVFormer & Voxels & 26.88 & 5.85 & 37.83 & 17.87 & 40.44 & 42.43 & 7.36 & 23.88 & 21.81 & 20.98 & 22.38 & 30.70 & 55.35 & 28.36 & 36.0 & 28.06 & 20.04 & 17.69  \\

        CTF-Occ & Voxels & 28.53 & 8.09 & 39.33 & 20.56 & 38.29 & 42.24 & 16.93 & 24.52 & 22.72 & 21.05 & 22.98 & 31.11 & 53.33 & 33.84 & 37.98 & 33.23 & 20.79 & 18.0 \\

        RenderOcc & Lidar & 23.93 & 5.69 &27.56 &14.36 &19.91 &20.56 &11.96 &12.42 &12.14 &14.34 &20.81 &18.94 & \textbf{68.85} &33.35 &42.01 &43.94 &17.36 &22.61  \\     
        RenderOcc & Voxels+Lidar & 26.11 & 4.84 & 31.72 & 10.72 & 27.67 & 26.45 & 13.87 & 18.2 & 17.67 & 17.84 & 21.19 & 23.25 & 63.2 & \textbf{36.42} & \textbf{46.21} & \textbf{44.26} & 19.58 & 20.72 \\ \hline
        
        TPVFormer & Voxels & 27.83 & 7.22 & 38.90 & 13.67 & 40.78 & 45.90 & 17.23 & 19.99 & 18.85 & 14.30 & \textbf{26.69} & \textbf{34.17} & 55.65 & 35.47 & 37.55 & 30.70 & 19.40 & 16.78 \\
        \rowcolor{Apricot!20!}
        \hspace{.3cm} w/ GaussRender & Voxels & 30.48 & 9.84 & 42.3 & 24.09 & 41.79 & 46.49 & 18.22 & 25.85 & 25.06 & 22.53 & 22.9 & 33.34 & 58.86 & 33.19 & 36.57 & 31.84 & 23.55 & 21.8 \\
        (gain) & & \textcolor{ForestGreen}{2.65} & \textcolor{ForestGreen}{2.62} & \textcolor{ForestGreen}{3.40} & \textcolor{ForestGreen}{10.42} & \textcolor{ForestGreen}{1.01} & \textcolor{ForestGreen}{0.59} & \textcolor{ForestGreen}{0.99} & \textcolor{ForestGreen}{5.86} & \textcolor{ForestGreen}{6.21} & \textcolor{ForestGreen}{8.23} & \textcolor{red}{-3.79} & \textcolor{red}{-0.83} & \textcolor{ForestGreen}{3.21} & \textcolor{red}{-2.28} & \textcolor{red}{-0.98} & \textcolor{ForestGreen}{1.14} & \textcolor{ForestGreen}{4.15} & \textcolor{ForestGreen}{5.02}
        \\ \hline

        SurroundOcc  & Voxels & 29.21 & 8.64 & 40.12 & 23.36 & 39.89 & 45.23 & 17.99 & 24.91 & 22.66 & 18.11 & 21.64 & 32.5 & 57.6 & 34.1 & 35.68 & 32.54 & 21.27 & 20.27 \\
        \rowcolor{Apricot!20!}
        \hspace{.3cm} w/ GaussRender & Voxels & 30.38 & 8.87 & 40.98 & 23.25 & 43.76 & 46.37 & 19.49 & 25.2 & 23.96 & 19.08 & 25.56 & 33.65 & 58.37 & 33.28 & 36.41 & 33.21 & 22.76 & 22.19 \\
        (gain) &  & \textcolor{ForestGreen}{1.17} & \textcolor{ForestGreen}{0.23} & \textcolor{ForestGreen}{0.86} & \textcolor{red}{-0.11} & \textcolor{ForestGreen}{3.87} & \textcolor{ForestGreen}{1.14} & \textcolor{ForestGreen}{1.50} & \textcolor{ForestGreen}{0.29} & \textcolor{ForestGreen}{1.30} & \textcolor{ForestGreen}{0.97} & \textcolor{ForestGreen}{3.92} & \textcolor{ForestGreen}{1.15} & \textcolor{ForestGreen}{0.77} & \textcolor{red}{-0.82} & \textcolor{ForestGreen}{0.73} & \textcolor{ForestGreen}{0.67} & \textcolor{ForestGreen}{1.49} & \textcolor{ForestGreen}{1.92} \\ \hline
        \end{tabular}
        }
    \caption{\textbf{Semantic voxel occupancy results on the \textbf{Occ3D-nuScenes} \citep{tian2023occ3d} validation set.} The best results are in bold. Training models with our module \method{} achieves state-of-the-art performance. Previous results are reported from \cite{huang2024gaussianformer2, pan2024renderocc}.}
    \label{tab:Occ3d-nuScenes}
\end{table*}

%% file: tables/kitti360_results.tex
\begin{table*}[t] %
    \small
    \setlength{\tabcolsep}{0.005\linewidth}  
    \centering
    \resizebox{\textwidth}{!}{
    \begin{tabular}{l|cc|cccccccccccccccccccc}
        \toprule
        Method &
        IoU &
        mIoU &
        \rotatebox{90}{\textcolor{ncar}{$\blacksquare$} car} &
        \rotatebox{90}{\textcolor{nbicycle}{$\blacksquare$} bicycle} &
        \rotatebox{90}{\textcolor{nmotor}{$\blacksquare$} motorcycle} & 
        \rotatebox{90}{\textcolor{ntruck}{$\blacksquare$} truck} &
        \rotatebox{90}{\textcolor{nother}{$\blacksquare$} other veh.} &
        \rotatebox{90}{\textcolor{npedestrian}{$\blacksquare$} person} &
        \rotatebox{90}{\textcolor{ndriveable}{$\blacksquare$} road} &
        \rotatebox{90}{\textcolor{nother}{$\blacksquare$} parking} &
        \rotatebox{90}{\textcolor{nsidewalk}{$\blacksquare$} sidewalk} &
        \rotatebox{90}{\textcolor{nother}{$\blacksquare$} other grnd.} &
        \rotatebox{90}{\textcolor{nmanmade}{$\blacksquare$} building} &
        \rotatebox{90}{\textcolor{nmanmade}{$\blacksquare$} fence} &
        \rotatebox{90}{\textcolor{nvegetation}{$\blacksquare$} vegetation} & 
        \rotatebox{90}{\textcolor{nterrain}{$\blacksquare$} terrain} &
        \rotatebox{90}{\textcolor{nmanmade}{$\blacksquare$} pole} &
        \rotatebox{90}{\textcolor{ntraffic}{$\blacksquare$} traf.-sign} &
        \rotatebox{90}{\textcolor{nmanmade}{$\blacksquare$} other struct.} &
        \rotatebox{90}{\textcolor{nother}{$\blacksquare$} other obj.} \\
        \midrule
        MonoScene * & 37.87 & 12.31 & 19.34 & 0.43 & 0.58 & 8.02 & 2.03 & 0.86 & 48.35 & 11.38 & 28.13 & 3.32 & 32.89 & 3.53 & 26.15 & 16.75 & 6.92 & 5.67 & 4.20 & 3.09 \\
        VoxFormer * & 38.76 & 11.91 & 17.84 & 1.16 & 0.89 & 4.56 & 2.06 & 1.63 & 47.01 & 9.67 & 27.21 & 2.89 & 31.18 & 4.97 & 28.99 & 14.69 & 6.51 & 6.92 & 3.79 & 2.43 \\
        OccFormer * & 40.27 & 13.81 & 22.58 & 0.66 & 0.26 & 9.89 & 3.82 & 2.77 & 54.30 & 13.44 & 31.53 & 3.55 & {36.42} & 4.80 & 31.00 & {\textbf{19.51}} & 7.77 & 8.51 & 6.95 & 4.60 \\
        \hline
        SurroundOcc & 38.51 & 13.08 & 21.31 & 0.0 & 0.0 & 6.05 & 4.29 & 0.0 & 53.88 & 12.56 & 30.89 & 2.57 & 34.93 & 3.59 & 29.03 & 16.98 & 5.61 & 6.66 & 4.39 & 2.62 \\
        \rowcolor{Apricot!20!}
        \hspace{.3cm} w/ GaussRender & 38.62 & 13.34 & 21.61 & 0.0 & 0.0 & 6.75 & 4.5 & 0.0 & 53.64 & 11.93 & 30.24 & 2.67 & 35.01 & 4.55 & 29.81 & 17.32 & 6.19 & 8.49 & 4.8 & 2.59 \\
        (gain) & \textcolor{ForestGreen}{0.11} & \textcolor{ForestGreen}{0.26} & \textcolor{ForestGreen}{0.30} & 0.0 & 0.0 & \textcolor{ForestGreen}{0.70} & \textcolor{ForestGreen}{0.21} & 0.0 & \textcolor{red}{-0.24} & \textcolor{red}{-0.63} & \textcolor{red}{-0.65} & \textcolor{ForestGreen}{0.10} & \textcolor{ForestGreen}{0.08} & \textcolor{ForestGreen}{0.96} & \textcolor{ForestGreen}{0.78} & \textcolor{ForestGreen}{0.34} & \textcolor{ForestGreen}{0.58} & \textcolor{ForestGreen}{1.83} & \textcolor{ForestGreen}{0.41} & \textcolor{red}{-0.03} \\  
        \hline 
        Symphonies (official checkpoint) & 43.40 & 17.82 & 26.86 & 4.21 & 4.92 & 14.19 & 7.67 & \textbf{16.79} & 57.31 & 13.60 & 35.25 & 4.58 & 39.20 & \textbf{7.96} & 34.23 & 19.20 & 8.22 & \textbf{16.79} & 6.03 & 6.03 \\
        \rowcolor{Apricot!20!}
        \hspace{.3cm} w/ GaussRender & \textbf{44.08} & \textbf{18.11} & \textbf{27.37} & 3.24 & \textbf{5.12} & \textbf{14.69} & \textbf{8.76} & 16.70 & \textbf{58.05} & \textbf{13.87} & \textbf{35.70} & \textbf{4.76} & \textbf{40.09} & 7.88 & \textbf{34.76} & 19.20 & \textbf{8.22} & 16.49 & \textbf{8.64 }& \textbf{6.50} \\
        (gain) &
        \textcolor{ForestGreen}{+0.68} & \textcolor{ForestGreen}{+0.29} & \textcolor{ForestGreen}{+0.51} & \textcolor{red}{-0.97} & \textcolor{ForestGreen}{+0.20} &
        \textcolor{ForestGreen}{+0.50} & \textcolor{ForestGreen}{+1.09} & \textcolor{red}{-0.09} & \textcolor{ForestGreen}{+0.74} & \textcolor{ForestGreen}{+0.27} &
        \textcolor{ForestGreen}{+0.45} & \textcolor{ForestGreen}{+0.18} & \textcolor{ForestGreen}{+0.89} & \textcolor{red}{-0.08} & \textcolor{ForestGreen}{+0.53} &
        \textcolor{gray}{0.00} & \textcolor{gray}{0.00} & \textcolor{red}{-0.30} & \textcolor{ForestGreen}{+2.61} & \textcolor{ForestGreen}{+0.47} \\
        \bottomrule
    \end{tabular}}
    \caption{\textbf{Semantic voxel occupancy results on the \textbf{SSCBench-KITTI360} \citep{li2024sscbench} test set.} The best results are in bold. Training models with our module \method{} achieves state-of-the-art performance. Previous results are reported from \cite{jiang2024symphonies}.}
    \label{tab:SSCBench-KITTI360}
    \end{table*}

%% file: fig/qualitative.tex
\begin{figure*}[t]
    \centering
    \begin{subfigure}{\textwidth}
        \centering
        \includegraphics[width=\linewidth]{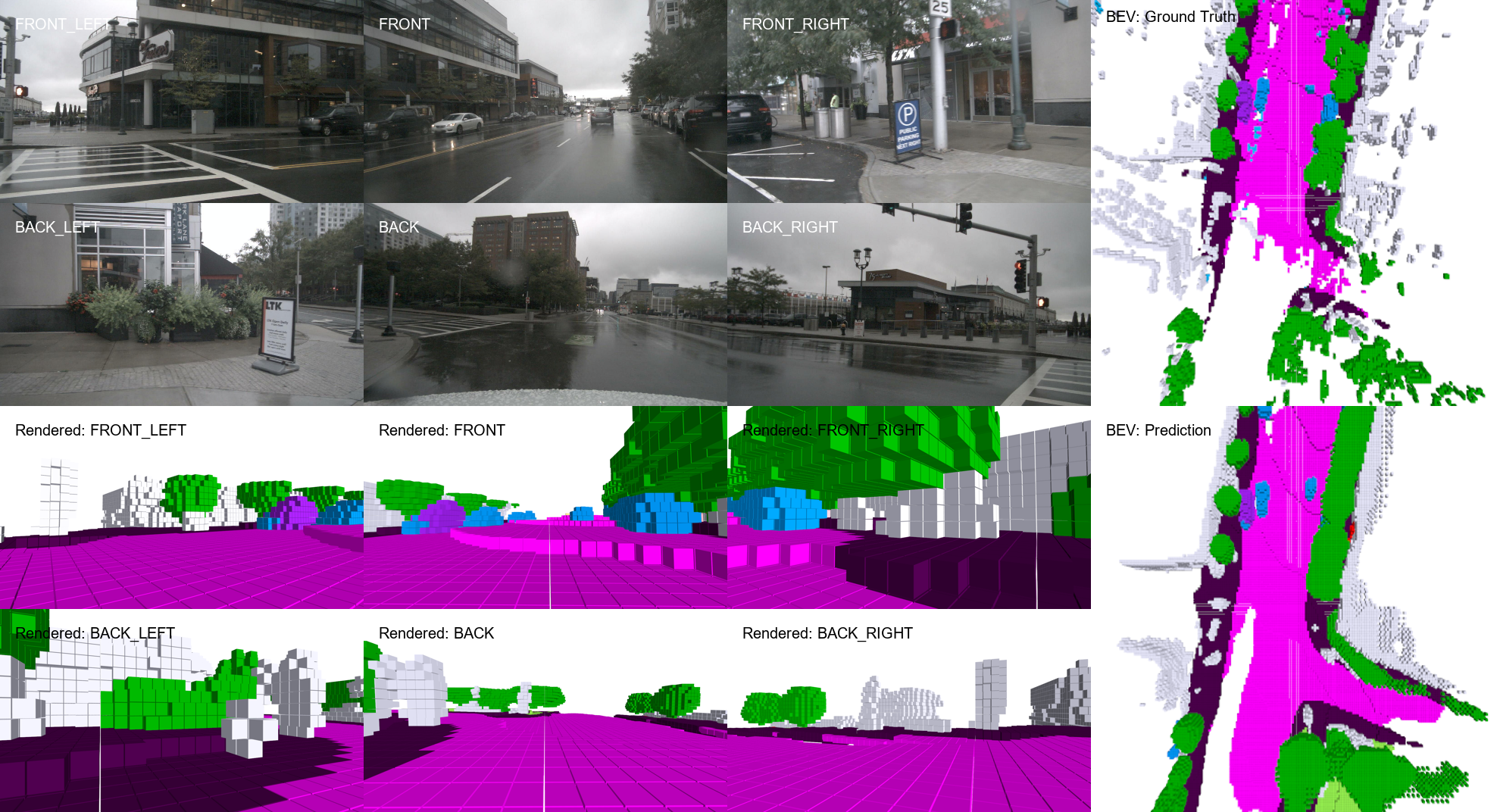}
    \end{subfigure}
    \hfill
    \vspace{0.2cm}
    \begin{subfigure}{\textwidth}
        \centering
        \includegraphics[width=\linewidth]{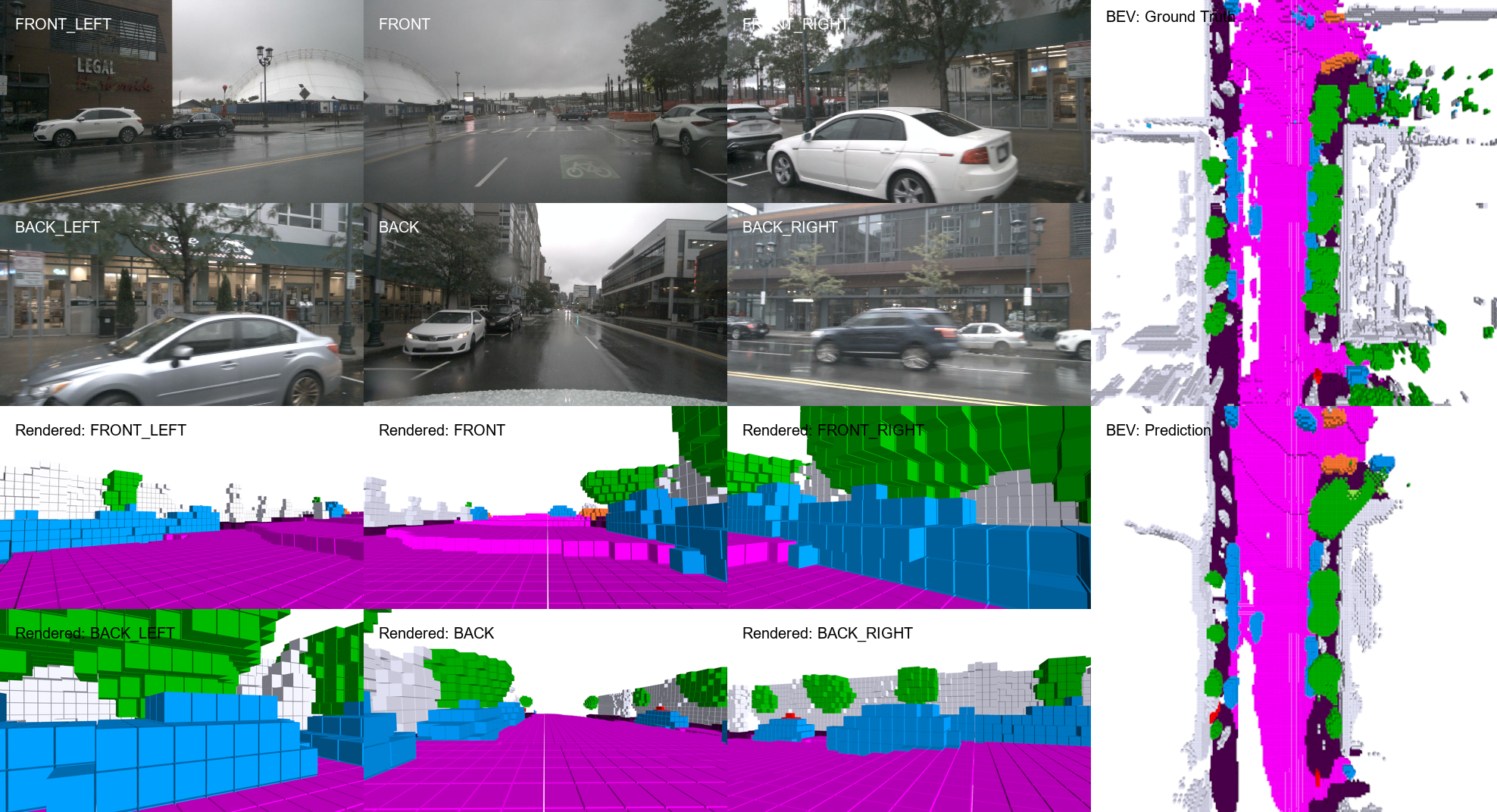}
    \end{subfigure}
    \begin{subfigure}{\textwidth}
        \centering
        \includegraphics[width=\linewidth]{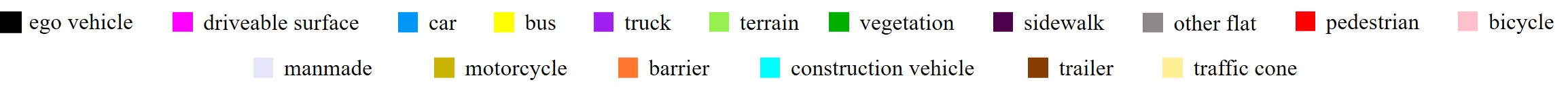}
    \end{subfigure}
    \caption{\textbf{Qualitative predictions of a SurroundOcc \cite{wei2023surroundocc} model trained with \method{} on the SurroundOcc-nuScenes \cite{wei2023surroundocc} dataset.} We display the six input camera images (top left), the rendered predictions (bottom left), the BeV ground-truth (top right) and BeV prediction (bottom left). The scene is randomly selected from the validation set and we show predictions at two different timesteps.}
    \label{fig:qualitative_0}
\end{figure*}

\begin{figure*}[t]
    \centering
    \begin{subfigure}{\textwidth}
        \centering
        \includegraphics[width=\linewidth]{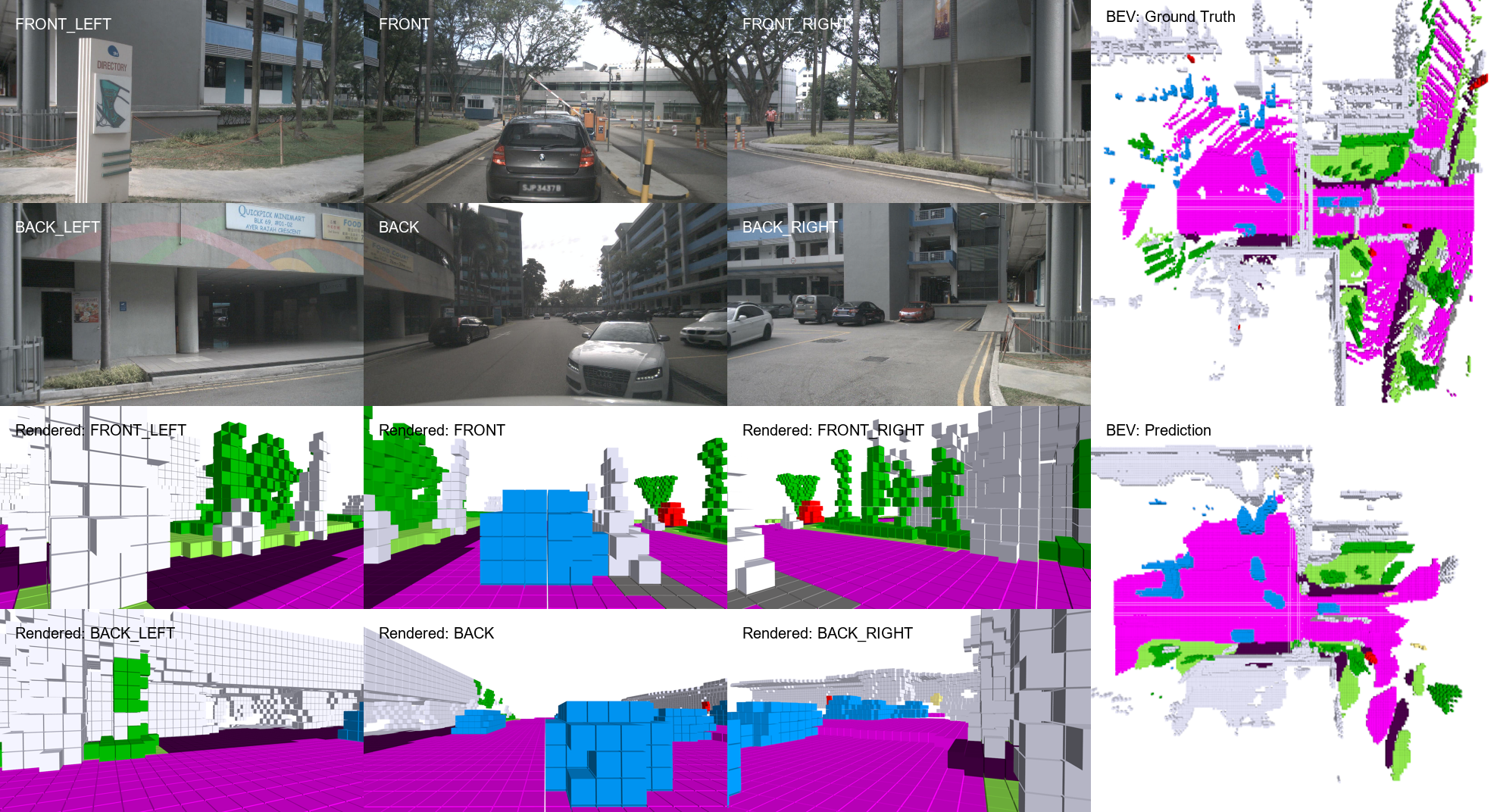}
    \end{subfigure}
    \hfill
    \vspace{0.2cm}
    \begin{subfigure}{\textwidth}
        \centering
        \includegraphics[width=\linewidth]{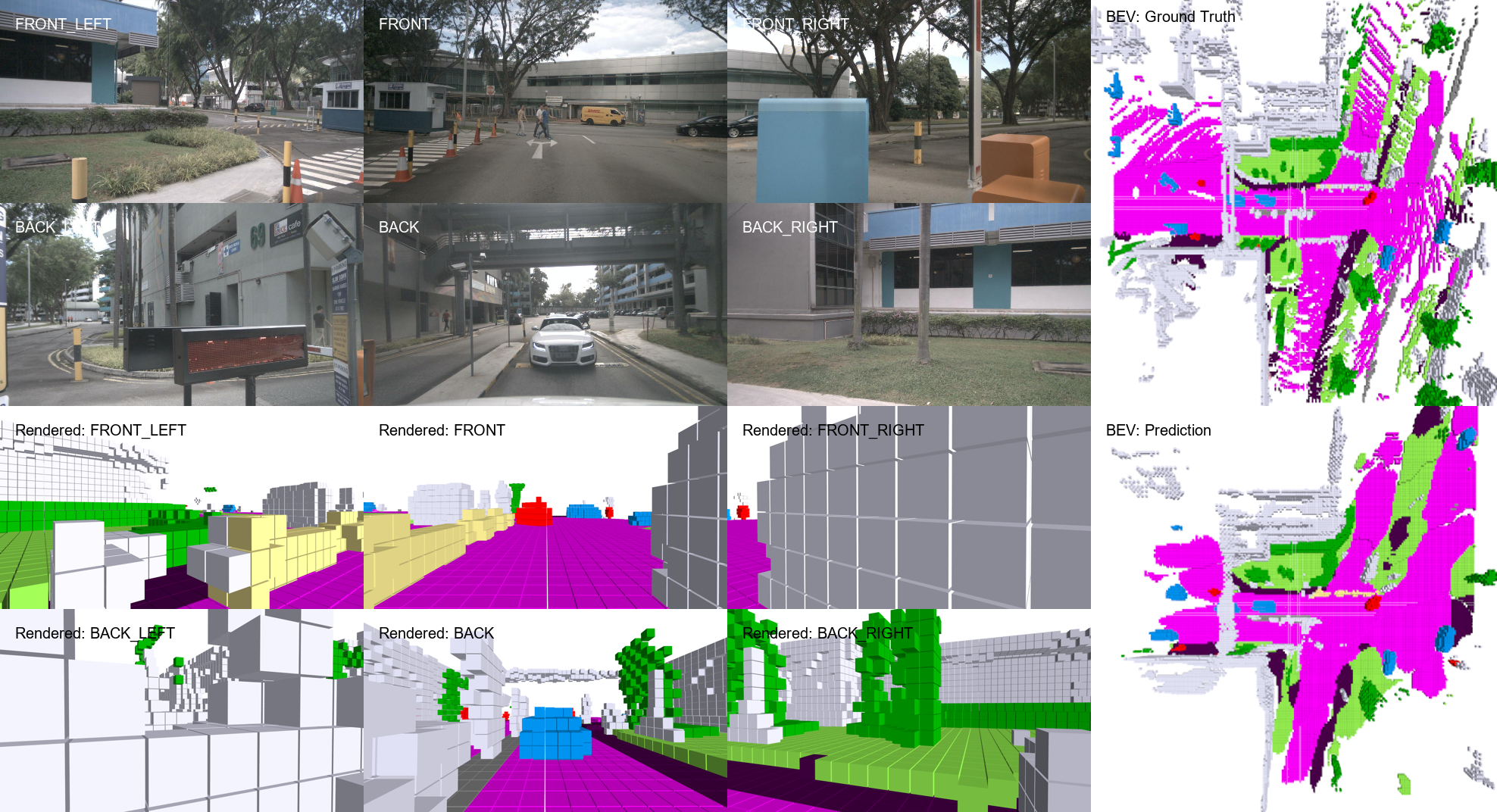}
    \end{subfigure}
    \begin{subfigure}{\textwidth}
        \centering
        \includegraphics[width=\linewidth]{fig/bar.jpg}
    \end{subfigure}
    \caption{\textbf{Qualitative predictions of a TPVFormer \cite{huang2023tpv} model trained with \method{} on the Occ3d-nuScenes \citep{tian2023occ3d} dataset.} We display the six input camera images (top left), the rendered predictions (bottom left), the BeV ground-truth (top right) and BeV prediction (bottom left). The scene is randomly selected from the validation set and we show predictions at two different timesteps.}
    \label{fig:qualitative_1}
\end{figure*}